\pdfoutput=1
\documentclass{article}

\PassOptionsToPackage{numbers, sort}{natbib}

 \usepackage[main, final]{neurips_2026}

\usepackage[utf8]{inputenc}
\usepackage[T1]{fontenc}
\usepackage{hyperref}
\usepackage{url}
\usepackage{booktabs}
\usepackage{amsfonts}
\usepackage{nicefrac}
\usepackage{microtype}
\usepackage{xcolor}

\usepackage[most]{tcolorbox}
\usepackage{microtype}
\usepackage{graphicx}
\usepackage{subfigure}
\usepackage{booktabs}
\usepackage{arydshln}
\usepackage{tikz}
\usetikzlibrary{bayesnet}
\usetikzlibrary{arrows}
\usepackage{amsmath}
\usepackage{amssymb}
\usepackage{mathtools}
\usepackage{amsthm}
\usepackage{pgfplots}

\usepackage{enumitem,multirow,adjustbox}

\usepackage{hyperref}
\usepackage{url}
\usepackage{xcolor}
\definecolor{darkblue}{rgb}{0, 0, 0.5}
\definecolor{beaublue}{rgb}{0.74, 0.83, 0.9}
\definecolor{gainsboro}{rgb}{0.86, 0.86, 0.86}
\definecolor{kleinblue}{rgb}{0,0.18,0.65}
\hypersetup{colorlinks=true,citecolor=kleinblue, linkcolor=kleinblue, urlcolor=kleinblue}

\usepackage{pifont}
\newtheorem{theorem}{Theorem}[section]
\newtheorem{proposition}[theorem]{Proposition}

\newtheorem{corollary}[theorem]{Corollary}
\newtheorem{definition}[theorem]{Definition}
\newtheorem{assumption}[theorem]{Assumption}

\usepackage{amsmath,amsfonts,bm}

\def\eqref#1{equation~\ref{#1}}

\def\1{\bm{1}}

\def\vx{{\bm{x}}}
\def\vy{{\bm{y}}}

\DeclareMathAlphabet{\mathsfit}{\encodingdefault}{\sfdefault}{m}{sl}
\SetMathAlphabet{\mathsfit}{bold}{\encodingdefault}{\sfdefault}{bx}{n}

\def\gD{{\mathcal{D}}}

\usepackage{enumitem,algorithm,algorithmic}

\usepackage{xspace}

\makeatletter
\newcommand*\rel@kern[1]{\kern#1\dimexpr\macc@kerna}
\newcommand*\widebar[1]{%
  \begingroup
  \def\mathaccent##1##2{%
    \rel@kern{0.8}%
    \overline{\rel@kern{-0.8}\macc@nucleus\rel@kern{0.2}}%
    \rel@kern{-0.2}%
  }%
  \macc@depth\@ne
  \let\math@bgroup\@empty \let\math@egroup\macc@set@skewchar
  \mathsurround\z@ \frozen@everymath{\mathgroup\macc@group\relax}%
  \macc@set@skewchar\relax
  \let\mathaccentV\macc@nested@a
  \macc@nested@a\relax111{#1}%
  \endgroup
}
\makeatother

\newenvironment{myquotation}{\setlength{\leftmargini}{0em}\quotation}{\endquotation}
\usepackage[hyperpageref]{backref}

\renewcommand*{\backrefalt}[4]{
  \ifcase #1 \relax
  \or
    (Cited on page #2)
  \else
    (Cited on pages #2)
  \fi
}

\definecolor{Gray}{gray}{0.9}

\graphicspath{{./fig/}}

\newtcolorbox[list inside=prompt,auto counter,number within=section]{prompt}[1][]{
    colbacktitle=black!80,
    colframe=black!80,
    coltitle=white,
    fontupper=\footnotesize,
    boxsep=5pt,
    left=0pt,
    right=0pt,
    top=0pt,
    bottom=0pt,
    boxrule=1pt,
    enhanced,
    breakable,
    skin first=enhanced,
    skin middle=enhanced,
    skin last=enhanced,
    #1,
}

\definecolor{light-purple}{RGB}{151,156,171}
\definecolor{blue-color}{RGB}{40,166,189}
\definecolor{pink-color}{RGB}{237,46,104}
\definecolor{dark-grey-color}{RGB}{79,91,102}

\usepackage{wrapfig}
\usepackage{bbm}
\usepackage{colortbl}
\definecolor{aliceblue}{RGB}{178, 217, 245}

\definecolor{babyblue}{RGB}{217, 239, 251}

\definecolor{babypink}{RGB}{251, 231, 230}
\definecolor{mygreen}{HTML}{3cb44b}
\definecolor{purple}{HTML}{7D2882}
\definecolor{darkred}{HTML}{B22222}
\definecolor{rosered}{HTML}{FF007F}
\definecolor{turquoise}{HTML}{00ced1}
\definecolor{darkturquoise}{HTML}{3c958e}

\newcommand{\oursfull}[0]{\texttt{Collaborative Multi-Agent Debate}\xspace}
\newcommand{\ours}[0]{\texttt{ColMAD}\xspace}
\newcommand{\copmad}[0]{\texttt{CopMAD}\xspace}
\newcommand{\cosmad}[0]{\texttt{CosMAD}\xspace}
\newcommand{\sa}[0]{\texttt{SA}\xspace}
\newcommand{\mad}[0]{\texttt{MAD}\xspace}
\newcommand{\real}[0]{\texttt{ReaLMistake}\xspace}
\newcommand{\mathw}[0]{\texttt{Math Word Problem Generation}\xspace}
\newcommand{\fine}[0]{\texttt{Fine-grained Fact Verification}\xspace}
\newcommand{\answer}[0]{\texttt{Answerability Classification}\xspace}

\usepackage{etoc}
\etocdepthtag.toc{mtchapter}
\etocsettagdepth{mtchapter}{subsection}
\etocsettagdepth{mtappendix}{none}

\usepackage{longtable}
\newcommand{\ccell}[1]{\multicolumn{1}{c}{#1}}

\newcommand{\mrtwo}[1]{\multirow{2}{*}{#1}}

\newcommand{\scb}[1]{\scalebox{0.8}[1]{#1}}

\title{When and Why Does Multi-Agent Debate Fail and Does It Really Underperform?}

\author{Yongqiang Chen$^{1,2}$\footnotemark[1]\hspace{0.45em},
Gang Niu$^{2}$,
James Cheng$^{1}$,
Bo Han$^{3,2}$,
Masashi Sugiyama$^{2,4}$
\vspace{1em}
\\
\textsuperscript{1}The Chinese University of Hong Kong, \hspace{0.5em}
\textsuperscript{2}RIKEN Center for Advanced Intelligence Project,\\
\textsuperscript{3}Hong Kong Baptist University, \hspace{0.5em}
\textsuperscript{4}The University of Tokyo\\
\texttt{\{yqchen,jcheng\}@cse.cuhk.edu.hk}, \hspace{0.5em}
\texttt{gang.niu.ml@gmail.com}\\
\texttt{bhanml@comp.hkbu.edu.hk}, \hspace{0.5em}
\texttt{sugi@k.u-tokyo.ac.jp}
}

\begin{document}

\maketitle
{\renewcommand{\thefootnote}{\fnsymbol{footnote}}%
\footnotetext[1]{Work done during an internship at RIKEN Center for Advanced Intelligence Project.}}

\begin{abstract}
Multi-agent debate (\mad) was proposed as a promising
approach for ensembling the wisdom of multiple large language
models (LLMs) to improve reasoning and provide effective
supervision to superhuman LLMs.
However, increasing empirical evidence suggests that \mad
may not outperform or even underperform
single-agent approaches (\sa), raising doubts about the
benefits of \mad.
In this work, we investigate this issue by analyzing the
incentive structures of popular \mad paradigms:
(i)~competitive \mad (\copmad) where agents compete by
holding opposing positions; (ii)~consensus-seeking \mad
(\cosmad) where agents are driven to seek consensus.
We show that both paradigms suffer from \textit{debate hacking}: \copmad reduces to a cheap-talk game, where agents produce misleading messages to win the game, while
\cosmad filters out informative disagreements for
premature consensus.
Consequently, agents in both \copmad and \cosmad fail to jointly resolve the ambiguity and seek the truth.
To this end, we introduce \ours,
a collaborative protocol that reframes \mad as a
non-zero-sum game to encourage agents to provide \textit{informative} while \textit{truthful} messages.
Through extensive benchmarking on challenging tasks such as error detection, we show that \ours consistently outperforms previous \mad protocols under multi-seed evaluation across heterogeneous LLM pairs. Under the same budgets, \ours effectively brings non-trivial improvements over \sa methods, implying that the protocol design is critical to realizing the potential of \mad. The code is available at \url{https://github.com/LFhase/ColMAD}.
\end{abstract}

\section{Introduction}
Along with the success of Large language models (LLMs) in solving tasks at various complexity levels~\citep{spark_AGI,o1,r1}, it has also drawn huge attention from the community to explore whether LLMs can exhibit collective intelligence and jointly resolve tasks with higher complexity.
Multi-agent debate (\mad) has been proposed to realize the intuition: multiple LLM agents hold different opinions for the same question, debate with each other to resolve the ambiguity, identify each other's oversight and refine the final answer~\citep{Chen2025HarnessingML,Feng2025WhenOL,Buhl2025AnAS}.
\mad has shown success in improving the reasoning of LLMs~\citep{du2023improving}, or even supervising powerful LLMs beyond human intelligence~\citep{irving2018ai}. It further motivates more sophisticated \mad approaches in improving the communication among the agents~\citep{yin-etal-2023-exchange,chan2023chateval}, and debating protocols~\citep{liang2023encouraging,chen-etal-2024-comm,khan2024debating,kenton2024on}.

However, increasing empirical observations also suggest that \mad may not outperform or even \textit{significantly underperform} single-agent (\sa) approaches such as chain-of-thought reasoning, despite the significantly larger token costs of \mad~\citep{wang-etal-2024-rethinking-bounds,Smit2023ShouldWB,Zhang2025IfMD,Yang2025RevisitingMD,kim2026capable}.
When \sa is scaled with self-consistency~\citep{wang2023selfconsistency}
under comparable token budgets, the gap widens further, raising a
natural question:
\begin{myquotation}\centering
\textit{Do we still need \mad? When and why does \mad underperform?}
\end{myquotation}

\begin{figure*}[t]
    \vspace{-0.15in}
	\centering
	\subfigure[Math problem]{
		\includegraphics[width=0.31\textwidth]{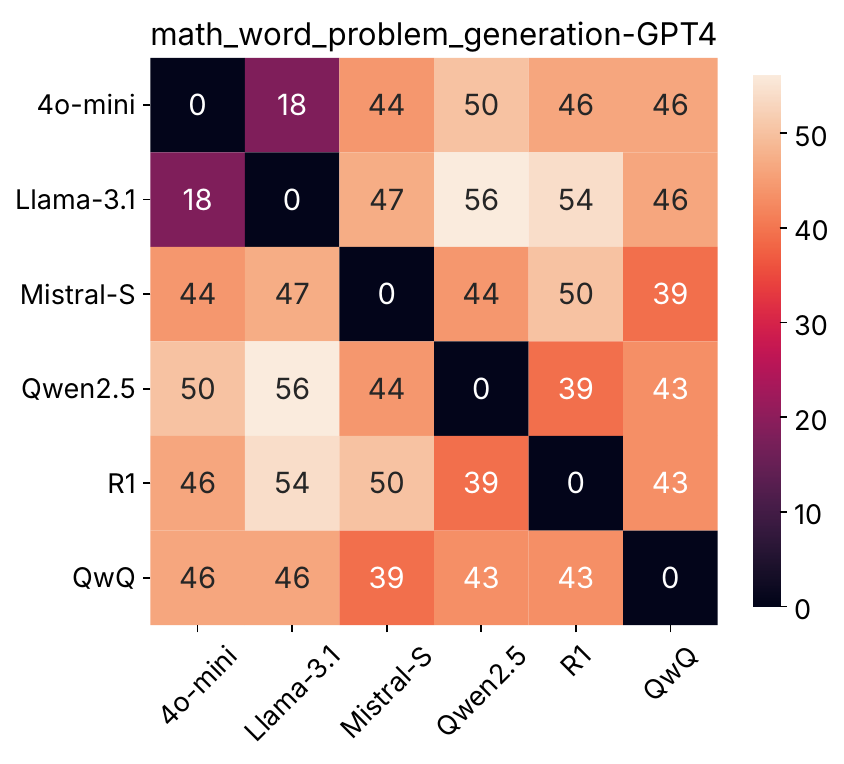}
	}
	\subfigure[Fact verification]{
		\includegraphics[width=0.31\textwidth]{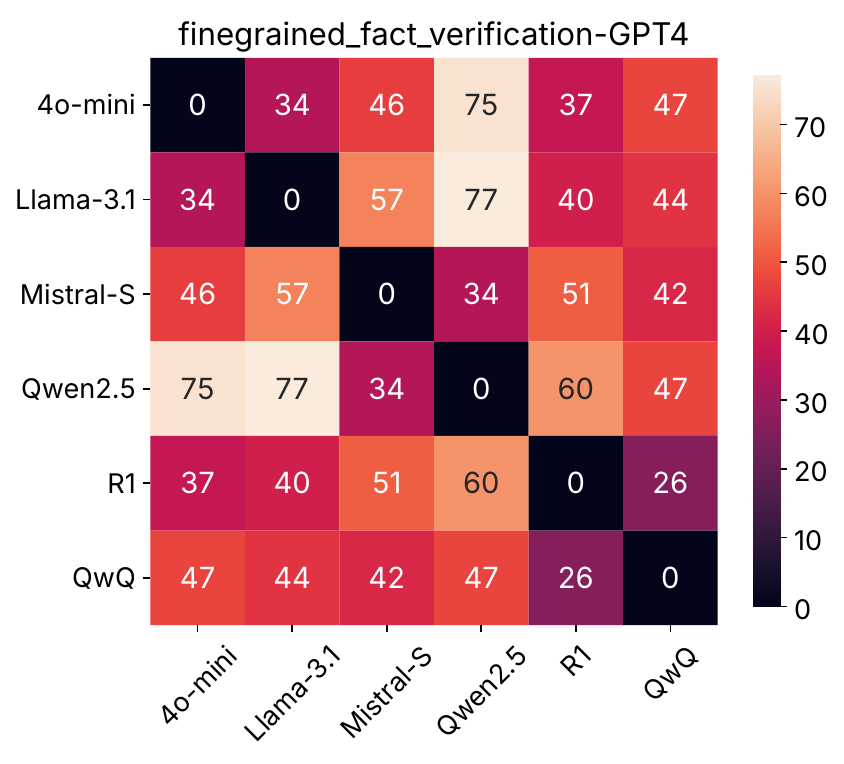}
	}
	\subfigure[Answerability]{
		\includegraphics[width=0.31\textwidth]{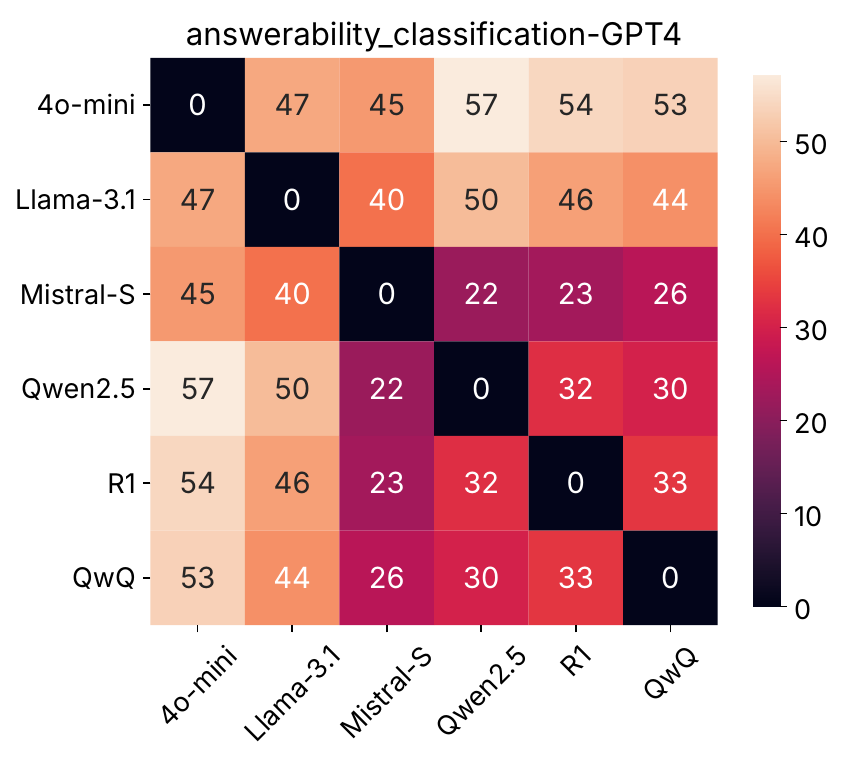}
	}
	\vspace{-0.05in}
	\caption{{Percent of reduced errors in detecting errors of \text{GPT-4}'s responses}. The numbers refer to the reduced errors following the oracle collaboration via Eq.~(\ref{eq:oracle_collab}). It can be found that different LLMs are less likely to make mistakes simultaneously. When incorporating LLMs with higher heterogeneity, such as those from different companies, the error reduction rates will be higher.}
	\label{fig:potential}
	\vspace{-0.15in}
\end{figure*}

In this work, we revisit the usefulness of \mad through their incentive structures. Specifically, existing \mad protocols can be categorized into (i)~competitive \mad (\copmad): agents are assigned different positions, and they debate with each other to convince the judge to take the respective position~\citep{irving2018ai,Buhl2025AnAS,liang2023encouraging,khan2024debating,kenton2024on}; (ii)~consensus-seeking \mad (\cosmad): agents may hold different positions and aim to reach to a consensus~\citep{du2023improving,liang2023encouraging,chan2023chateval,yin-etal-2023-exchange,chen-etal-2024-reconcile}.
We examine the effectiveness of \mad through the task of error detection \real~\citep{realmistake}, where agents need to find whether there is a mistake in a given LLM response.
Since LLMs differ in their knowledge and error tendencies~\citep{kim2025correlated}, different LLMs may provide \textit{complementary signals for error detection}, and jointly considering perspectives from different LLMs becomes more critical.
Although MAD has a great potential in error detection (see Fig.~\ref{fig:potential}), \copmad falls well below a single agent (see Fig.~\ref{fig:pitfall_mad}), and \cosmad improves over the single agents only when the two debaters have comparable capability.

\begin{figure}[!t]
    \vspace{-0.15in}
	\centering
	\subfigure[Illustration of \mad on error detection]{
		\includegraphics[width=0.6\textwidth]{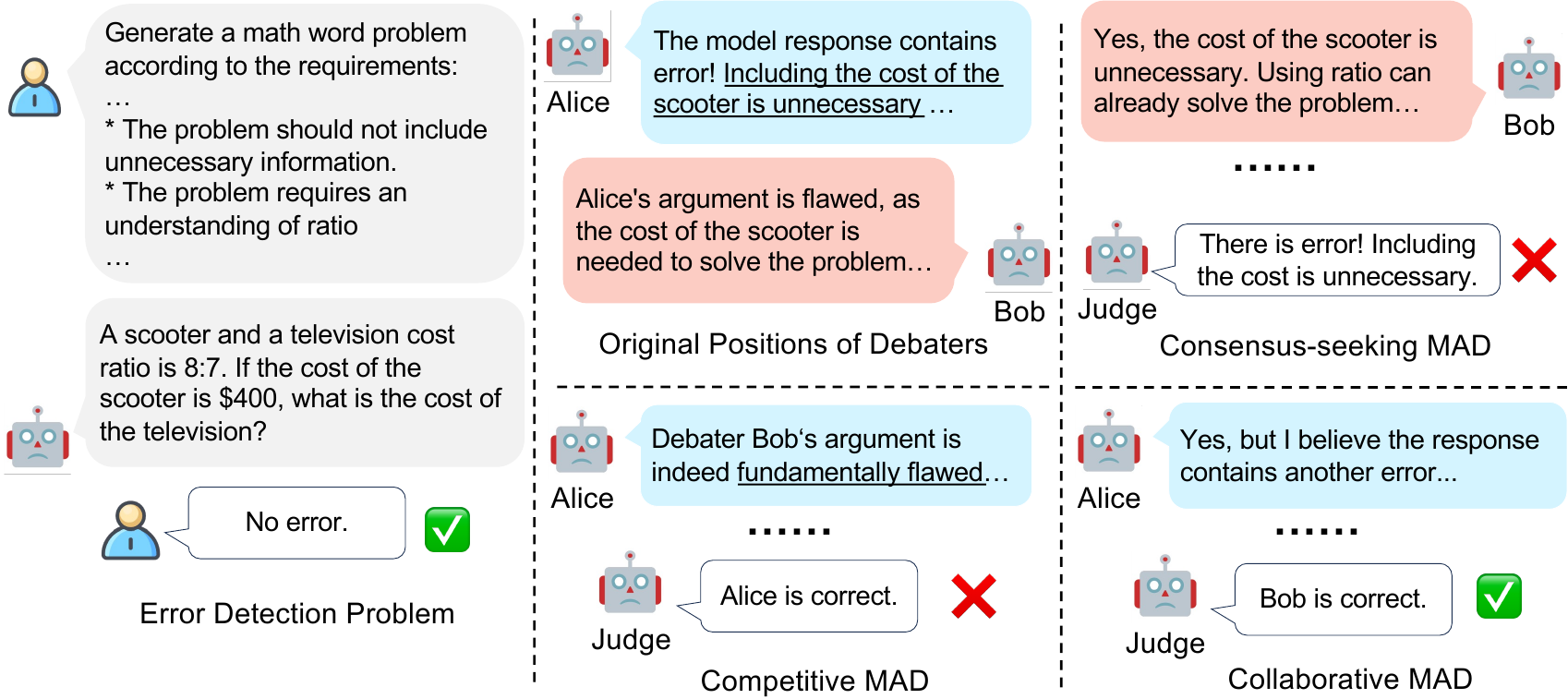}
	}
	\subfigure[Cost versus gain]{
		\includegraphics[width=0.30\textwidth]{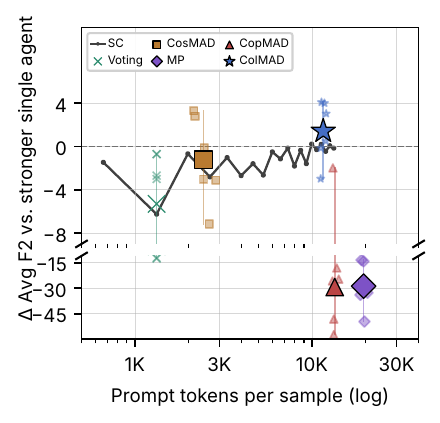}
	}
	\vspace{-0.05in}
	\caption{Comparison of different \mad protocols on error detection. (a)~Given a task of checking whether an LLM-generated math problem meets all requirements, both \copmad and \cosmad exhibit \textit{debate hacking}: \copmad debaters use overconfident rhetoric (``\textit{fundamentally flawed}'') to mislead the judge, while \cosmad debaters echo each other's opinion without challenge. \ours encourages debaters to complement missing evidence, enabling the judge to reach the correct verdict. (b)~Change in average F2 over the stronger single agent against prompt tokens per sample on the six pairs of LLMs from Table~\ref{tab:gpt4-results}. \ours is the only protocol whose mean is above the single agent and its worst pair is the mildest; \cosmad is cheap but below zero on average; \copmad loses 14 to 45 points on four of the six pairs and MP 24 and 40 on the two pairs with a reasoning model.}
	\label{fig:comad_illustration}
	\vspace{-0.15in}
\end{figure}

Interestingly, \mad exhibits \textit{debate hacking} behaviors (see Fig.~\ref{fig:comad_illustration}).
In \copmad, debaters tend to misinterpret the task requirements and present claims in an overconfident tone to mislead the judge agent.
While in \cosmad, debaters tend to neglect critical disagreements in order to reach consensus.
In other words, \copmad debaters will try every possibility to persuade the judge to \textit{win the game}, instead of providing \textit{honest and evidential statements}. Consequently, \copmad can even underperform single-agent methods.
Through a simple game-theoretic modeling of the two \mad protocols, we relate debate hacking to \textit{goal misalignment} in the protocol design: the incentive to \textit{win the game} in \copmad admits equilibria where the messages carry no information about the answer, so competitive debate guarantees no improvement in the worst case, while the incentive to \textit{agree with each other} in \cosmad discards informative disagreements.
In contrast, truth-seeking debate requires agents to provide  \textit{informative} and \textit{truthful} messages.

Motivated by our theoretical analysis, we propose a new \mad protocol called \oursfull (\ours) that reframes \mad as a non-zero-sum game for truth-seeking: instead of prompting the agents to win the game or merely find a consensus, \ours encourages agents to find new truthful information that supports their positions and to agree on correct points raised by the other debater, so that the judge can make a more informative decision from the transcripts (Proposition~\ref{thm:col}).
Empirically, on the three error-detection tasks of \real~\citep{realmistake} with six heterogeneous LLM pairs, \ours consistently outperforms both \copmad and \cosmad (Sec.~\ref{sec:exp}). Multi-seed evaluations also confirm the statistical significance of \ours over \copmad, \cosmad and \sa.
Extended comparisons on mathematical reasoning and safety alignment~\citep{Yang2025RevisitingMD} are consistent with this picture, and under the same token budgets \ours also outperforms \sa scaled with self-consistency and with voting: proper protocol design is what unlocks the potential of \mad.

\section{Revisiting Multi-Agent Debate in Error Detection}
\label{sec:method}

In this section, we revisit the effectiveness of \mad via error detection.
Without loss of generality, we consider the following debate setting with two agents $\mathrm{A}$ and $\mathrm{B}$ and a judge $J$.

\begin{definition}[Basic \mad setup]\label{def:basic_mad}
    Given a question $Q$ with a binary label $Y\in\{0,1\}$, two debaters $\mathrm{A}$ and $\mathrm{B}$ with different beliefs debate and produce messages $M=(m_A,m_B)$. Without loss of generality, $\mathrm{A}$ believes $Y_A=1$ and $\mathrm{B}$ believes $Y_B=0$. A judge $J$ gives predictions $Y_J=J(X_0,M)$ based on $X_0=(x_A,x_B)$ and $M$. We denote messages before round $t$ as $M^{(t-1)}$.
\end{definition}

We examine the potential of \mad in the task of error detection via the \real benchmark~\citep{realmistake}.
It contains $3$ objective error detection tasks: (i) math word problem generation (Math problem) that requires LLMs to generate math problems satisfying requirements; (ii) fine-grained fact verification (Fact verification) that requires LLMs to verify claims given fine-grained evidence; (iii) answerability classification (Answerability) that requires LLMs to leverage their commonsense knowledge to examine whether a question is answerable.
As LLMs show different error tendencies~\citep{kim2025correlated} and single LLM can hardly tell the errors in LLM responses when without no reliable external feedback~\citep{kamoi2024self-correction}, it is natural to jointly incorporate opinions from different LLMs. Crucially, \copmad is considered to be a promising scalable oversight approach to detect errors of LLM responses~\citep{irving2018ai,khan2024debating,kenton2024on}.

\subsection{Potential and Pitfalls of Multi-Agent Debate}
\label{sec:mad_potential}

With \real, we first consider the scenario where we can perfectly ensemble the knowledge of different LLMs:
Given Definition~\ref{def:basic_mad}, intuitively, if the elicited messages $M=(m_A,m_B)$ bring additional information to the judge, i.e., $I(Y;M|X_0)>0$ where $I(\cdot;\cdot)$ denotes the mutual information, the judge can make a more well-informed decision about $Y$ given $M$.
The complementary information lies in cases where the predictions of $\mathrm{A}$ differ from those of $\mathrm{B}$.
If both agents can provide \textit{sufficient justifications} and are more persuasive when debating for the \textit{correct} answer, the judge can be convinced to take the correct answer whenever either agent is correct.
Given a task associated with $n$ question-label pairs $\{q^{(i)},\vy^{(i)}\}_{i=1}^n$, and the predictions by $A$ and $B$ as $\vy^{(i)}_A$ and $\vy^{(i)}_B$ respectively, the reduction of errors from the oracle collaboration is
\begin{equation}\label{eq:oracle_collab}
	\min_{K\in\{A,B\}}\sum_{i=1}^n\mathbbm{1}[\hat{\vy}_K^{(i)}\neq\vy^{(i)}]-\sum_{i}\mathbbm{1}[\hat{\vy}_J^{(i)}\neq\vy^{(i)}],
\end{equation}
where $\mathbbm{1}[\cdot]$ is the indicator function that equals $1$ if the condition inside holds true and $0$ otherwise, $\vy^{(i)}_K$ denotes the initial prediction of agent $K\in\{A,B\}$ on the $i$-th sample.
Then, we plot the error reductions for the six LLMs used in our experiments in Fig.~\ref{fig:potential}. Different LLMs tend not to make the same mistakes: across all pairs and tasks, the oracle collaboration reduces the errors by 18\% to 77\%. This demonstrates the substantial untapped potential of ensembling opinions from different LLMs. Appendix~\ref{appdx:potential} extends the analysis to more LLMs and to the responses of Llama-2.
This demonstrates the substantial untapped potential of ensembling opinions from different LLMs.

\begin{figure*}[t]
    \vspace{-0.15in}
    \centering
    \subfigure[4o-mini + Llama-3.1]{
        \includegraphics[width=0.31\textwidth]{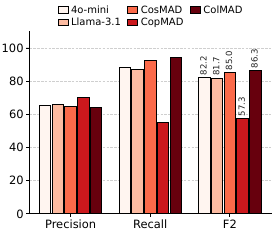}
    }
    \subfigure[4o-mini + R1]{
        \includegraphics[width=0.31\textwidth]{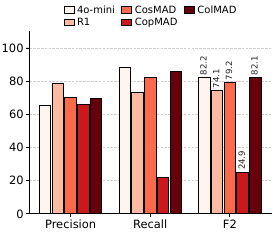}
    }
    \subfigure[Llama-3.1 + R1]{
        \includegraphics[width=0.31\textwidth]{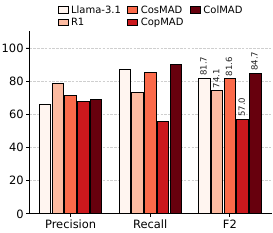}
    }
    \vspace{-0.05in}
    \caption{Precision, recall and F2 of \sa and \mad protocols on three debater pairs. Debater\#1 is listed first and also serves as the judge where applicable. \copmad falls far below the stronger \sa on all pairs. \cosmad exceeds both \sa in (a) and does not improve over the stronger one in (b) and (c). \ours achieves the highest F2 in (a) and (c) and stays at the level of the stronger \sa in (b).}
    \label{fig:pitfall_mad}
    \vspace{-0.15in}
\end{figure*}

Then, let us look into the practical \mad realizations. According to the incentive structures, existing \mad methods can be categorized into two: \copmad (i.e., Competitive \mad) where LLM agents competitively debate with each other to convince the judge of the respective assigned answers~\citep{khan2024debating}; and \cosmad (i.e., Consensus-seeking \mad) where LLM agents seek consensus~\citep{du2023improving}.
In Fig.~\ref{fig:pitfall_mad}, we conduct an initial experiment with \copmad and \cosmad adapted from previous \copmad~\citep{khan2024debating,kenton2024on} and \cosmad~\citep{du2023improving}, respectively.
Specifically, we consider three debater pairs formed by \texttt{GPT4o-mini} (4o-mini in short), \texttt{Llama-3.1-70B} (Llama-3.1 in short) and \texttt{DeepSeek-R1} (R1 in short).
We report the precision, recall, and F2 score, which is an instantiation of the $F_\beta$ score~\citep{fbeta_score}:
$
F_\beta=(1+\beta^2)\cdot\text{precision}\cdot\text{recall}/(\beta^2\cdot \text{precision}+\text{recall}),
$
with $\beta$ as $2$. We focus on the F2 score as it emphasizes recall, i.e., whether all errors are detected, which is critical to error detection.

From the results, \copmad and \cosmad show different failure patterns. \copmad cuts the recall sharply and falls far below the stronger single agent on all three pairs. \cosmad is a strong baseline when the two debaters have comparable F2, but no longer improves over the stronger single agent when their F2 differs substantially.
When inspecting the behaviors, \copmad and \cosmad show different patterns:
\cosmad achieves performance close to the \textit{average} of the two agents rather than improving upon either. When a strong agent collaborates with a weaker one, the strong agent abandons its own correct judgment in favor of premature consensus, suppressing critical disagreements.

\subsection{Why Do Existing Protocols Fail?}
\label{sec:mad_theory}
From the empirical results, \copmad exhibits the failure widely observed in the literature~\citep{wang-etal-2024-rethinking-bounds,Smit2023ShouldWB,Zhang2025IfMD,Yang2025RevisitingMD}, and \cosmad improves over \sa only when the debaters have comparable F2, despite the large potential shown in Fig.~\ref{fig:potential}.
To analyze the behaviors of debaters, we consider the following assumption that the final answer by the judge largely depends on what information is elicited through debating.

\begin{assumption}\label{assump:optimal_judge}
	Denoting the label as $Y\in\{0,1\}$ with prior $\pi\in(0,1)$, the judge derives the answer based on $X_0=(x_A,x_B)$ using the log-likelihood ratio (LLR): $\Lambda_0(X_0)=\log\frac{p(X_0|Y=1)}{p(X_0|Y=0)}+\log\frac{\pi}{1-\pi}.$
	Each debate message contributes $l_i(m_i;X_0)=\log\frac{p(m_i|Y=1,X_0)}{p(m_i|Y=0,X_0)},\ i\in\{A,B\}.$
    With debate, the total LLR is $\Lambda=\Lambda_0(X_0)+l_A(m_A;X_0)+l_B(m_B;X_0).$
\end{assumption}
Intuitively, Assumption~\ref{assump:optimal_judge} formalizes that debate can help the judge only when messages carry information beyond $X_0$, i.e., $I(Y;M\mid X_0)>0$. Otherwise $\Lambda(X_0,M)=\Lambda_0(X_0)$ almost surely, where $\Lambda(X_0,Z)=\log\frac{P(Y=1\mid X_0,Z)}{P(Y=0\mid X_0,Z)}$ denotes the posterior log-odds given $(X_0,Z)$, and the transcript does not change the judge's decision.
In \copmad~\citep{khan2024debating,kenton2024on}, debaters are assigned opposing positions, and each aims to convince the judge.
\begin{definition}[Competitive \mad]\label{def:copmad}
	Debaters $\mathrm{A}$ and $\mathrm{B}$ have opposed utilities:
	$u_A(m_A)=P(Y_J=Y_A\mid m_A,X_0,M^{(t-1)}),\  u_B(m_B)=P(Y_J=Y_B\mid m_B,X_0,M^{(t-1)}).$
    The risk is $V_\mathrm{cop} = \min_J \max_{e \in \mathcal{E}_\mathrm{cop}(J)} P(J(X_0, M_e) \neq Y)$, where $\mathcal{E}_\mathrm{cop}(J)$ is the set of Nash equilibria of the A-B subgame in convincing $J$, $M_e$ is the debating transcript under $e$, and the inner maximum selects the equilibrium that is worst for the judge.
\end{definition}

\copmad creates a \textit{zero-sum} incentive: each debater benefits from convincing the judge, \textit{regardless of correctness}. Essentially, \copmad naturally extends the typical cheap talk game~\citep{crawford1982strategic}, where debaters $\mathrm{A}$ and $\mathrm{B}$ can transmit costless messages to each other and to the judge $J$.
Since each debater's utility depends only on the judge's verdict, which is a function of $(X_0,M)$, the subgame admits a \textit{babbling equilibrium} for any judge, in which $M\perp Y\mid X_0$ and the messages are uninformative.

\begin{figure*}[t]
    \vspace{-0.15in}
    \centering
	\subfigure[Illustration of debate hacking]{
		\includegraphics[width=0.675\textwidth]{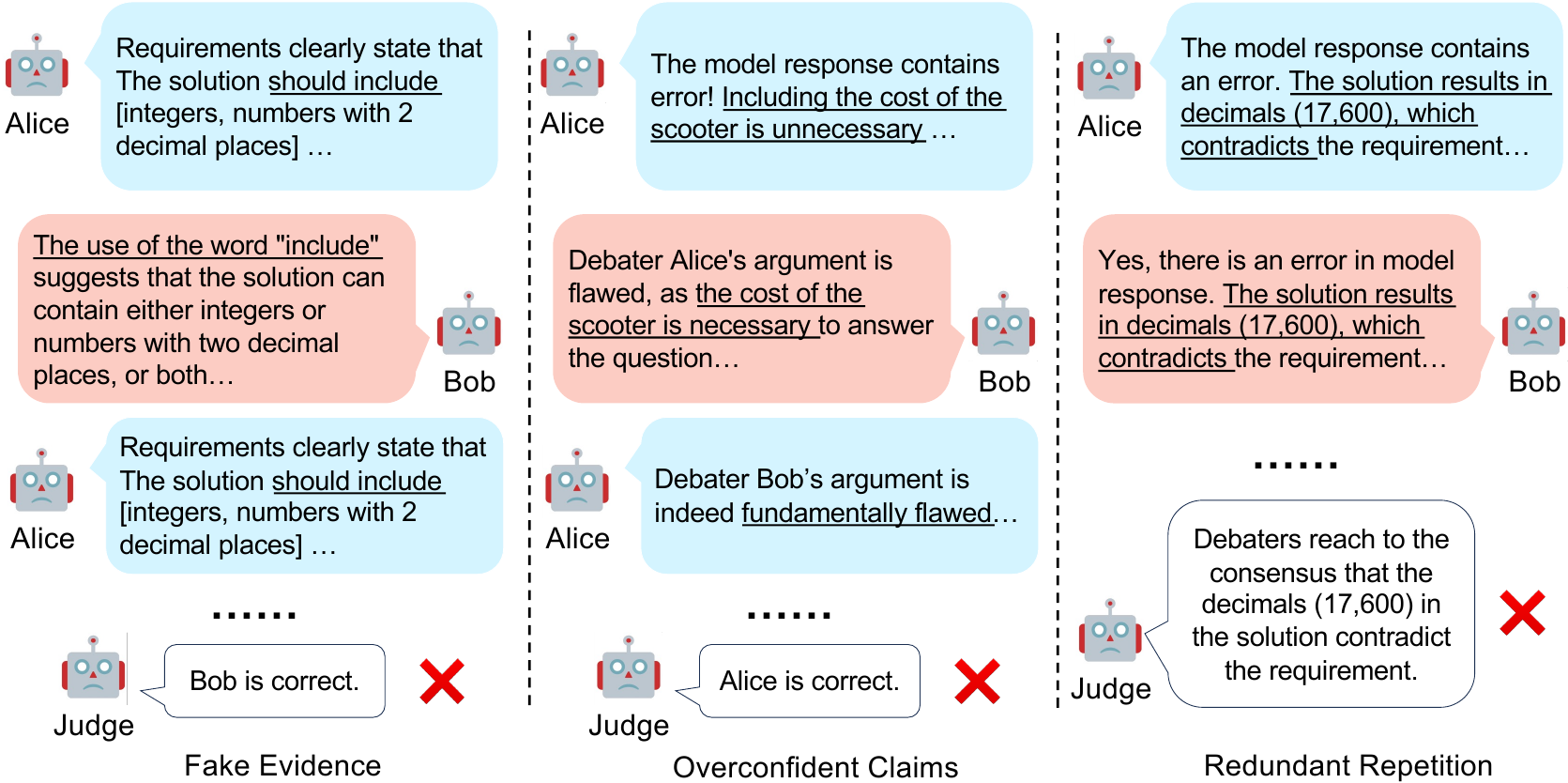}
	}
	\subfigure[Behavior analysis]{
		\includegraphics[width=0.285\textwidth]{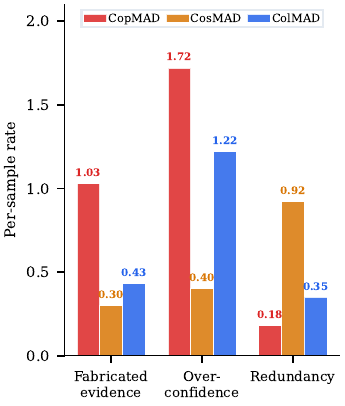}
	}
    \vspace{-0.05in}
    \caption{Debate hacking in existing \mad protocols. (a)~Three failure behaviors observed in debate transcripts: (i) \textit{Fake evidence} that dishonest debaters misinterpret the requirements of the task; (ii) \textit{Overconfident claims} that dishonest debaters use an overconfident tone to mislead the judge; and (iii) \textit{redundancy}, where debaters
    produce near-verbatim copies of each other's
    arguments instead of independent verification. The first two are characteristic of
    \copmad; the third is characteristic of \cosmad.
    (b)~Rubric-based behavioral analysis. \copmad is dominated by fabricated evidence and overconfidence, while
    \cosmad is dominated by redundancy. \ours
    reduces all three behaviors while maintaining truth-seeking competition.}
    \label{fig:debate_hacking}
    \vspace{-0.15in}
\end{figure*}

\begin{proposition}[Pitfalls of competitive debating]\label{thm:pitfal_debating}
	Assuming bounded LLR $|l_i|\leq L_i<\infty$, $i\in\{A,B\}$, let $R(Z)$ denote the Bayes risk of the judge $J$ when making decisions based on $Z$, and denote $R_0=R(X_0)$. Then $V_\mathrm{cop}=R_0$.
\end{proposition}
The proof is in Appendix~\ref{proof:pitfal_debating}.
Proposition~\ref{thm:pitfal_debating} is a worst-case statement: under the babbling equilibrium, $I(Y;M\mid X_0)=0$ and the Bayes-optimal judge gains nothing from the transcript, so the competitive incentive alone guarantees no improvement over $R_0$. This is compatible with \citet{khan2024debating} and \citet{kenton2024on}, who report settings where competitive debate is informative and supports scalable oversight, i.e., an informative equilibrium is reached. In practice, real judges are not Bayes-optimal and are susceptible to debate hacking, which can place \copmad \textit{below} the \sa baseline~\citep{Zhang2025IfMD}. On error detection, the fabricated evidence and overconfident claims in Fig.~\ref{fig:debate_hacking} are consistent with debaters reaching unfavorable equilibrium.
In the meantime, \citet{khan2024debating} and \citet{kenton2024on} report settings where competitive debate is informative and supports scalable oversight, i.e., an informative equilibrium is reached. Proposition~\ref{thm:pitfal_debating} is compatible with these findings: it shows that such gains are not guaranteed by the competitive incentive. On error detection, the fabricated evidence and overconfident claims in Fig.~\ref{fig:debate_hacking} are consistent with debaters reaching the unfavorable equilibria.
Let us then discuss \cosmad~\citep{du2023improving,chen-etal-2024-reconcile}, where agents revise toward agreement.
\begin{definition}[Consensus-seeking \mad]\label{def:cosmad}
Debaters $\mathrm{A}$ and $\mathrm{B}$ share the same utilities towards reaching consensus:
$u_A(m_A) = u_B(m_B) = P(Y_A = Y_B \mid
X_0, M^{(t-1)})$.
The risk is
$V_\mathrm{cos} = \min_J \min_{e \in
\mathcal{E}_\mathrm{cos}(J)}
P(J(X_0, M_e) \neq Y)$, where $M_e=\Phi_\cap(M)$ is the debating transcripts given by the optimal Nash equilibrium $e \in \mathcal{E}_\mathrm{cos}(J)$ of the A-B subgame in reaching consensus.
\end{definition}

Unlike \copmad, agents are not incentivized to persuade, but are implicitly incentivized to \textit{agree}. Essentially, the incentive creates a gate that filters out informative disagreements.

\begin{proposition}[Pitfalls of consensus-seeking debate]
\label{thm:cosmad}
Given the same setting as in Proposition~\ref{thm:pitfal_debating}, in \cosmad, we have
$R(X_0, \Phi_\cap(M)) \geq R(X_0, M)$, with strict inequality when
$P\big(\Lambda(X_0,\Phi_\cap(M))\,\Lambda(X_0,M)<0\big)>0$, i.e., when the disagreements discarded by the consensus process would overturn the judge's decision with positive probability.
\end{proposition}

The proof is in Appendix~\ref{proof:cosmad}. Proposition~\ref{thm:cosmad} explains the empirical observation about \cosmad that \cosmad reduces information not through fabrication, but through suppression of disagreements.

\subsection{Collaborative Multi-Agent Debate}
\label{sec:col_method}

Both protocols fail for the misaligned incentive with respect to truth-seeking.
To mitigate the issue, we introduce \oursfull (\ours) to encourage both conditions.
\begin{definition}[Collaborative \mad]\label{def:colmad}
Debaters $\mathrm{A}$ and $\mathrm{B}$ have \textit{truth-seeking} utilities: $u_A(m_A)=u_B(m_B)=I(Y;m_i\mid X_0,M^{(t-1)}),$
i.e., both debaters aim to reduce uncertainty about $Y$ by complementing missing information. The risk is $V_\mathrm{col}=\min_J\min_{e\in\mathcal{E}_\mathrm{col}(J)}P(J(X_0,M_e)\neq Y)$, where $M_e$ is the debating transcripts given by the Nash equilibrium $e\in\mathcal{E}_\mathrm{col}(J)$ of the A-B subgame in collaborative seeking the truth.
\end{definition}
Essentially, \ours turns the \mad into a \textit{non-zero-sum game} where agents are incentivized to \textit{collaborate}: surface missing evidence, verify claims against context, identify weaknesses in both sides' reasoning, and enable the judge to make a more informative decision.

\begin{proposition}[Collaborative debate]\label{thm:col}
Under the same setting, $V_\mathrm{col}\leq R(X_0)=V_\mathrm{cop}$, with strict inequality when $I(Y;M_e|X_0)>0$.
\end{proposition}
The proof of Proposition~\ref{thm:col} is given in Appendix~\ref{proof:col}.
Intuitively, if the debater agents can provide additional information about the question, e.g., pointing out the missing information from the reasoning of the other agent, the collaborative scheme is provably better than the competition.
The above three propositions establish a clear ordering:

\begin{corollary}[Ordering of \mad protocols]\label{cor:ordering}
$V_\mathrm{col}\leq V_\mathrm{cos}\leq R(X_0)=V_\mathrm{cop}.$
\end{corollary}

Each gap has a clear cause: $V_\mathrm{cop}=R(X_0)$ because competitive incentives produce uninformative messages; $V_\mathrm{cos}\leq R(X_0)$ because consensus retains \textit{some} information but discards disagreements; $V_\mathrm{col}\leq V_\mathrm{cos}$ because collaborative debate preserves the full transcript including disagreements.

\textbf{Theory-inspired protocol design.} Motivated by the theoretical analysis, we design \ours protocol to explicitly encourage the respective truthfulness and informativeness conditions: (i)~\textit{evidence verification} via a quote-based system with exact-match checking, and \textit{self-auditing} where debaters identify potential failure modes in their own claims; (ii)~a \textit{collaborative objective} that asks debaters to complement missing points, and \textit{confidence calibration} that helps the judge weight contributions. A detailed mapping is in Appendix~\ref{appdx:theory_impl_map}.
More details are given in Appendix~\ref{appdx:method}.

\section{Related Work}
\label{sec:related}
In this section, we discuss the related work and background of \mad.

\textbf{Multi-Agent Debate.}
\mad aims to imitate the cooperation of humans towards building a society of AI~\citep{Minsky1987TheSO}. Recently, \mad has gained significant attention as one of the promising approaches to scale up the test-time computation for enhancing reasoning and alignment~\citep{ai_safety_debate,du2023improving,khan2024debating,kenton2024on}. A full categorization of prior \mad approaches is given in Appendix~\ref{appdx:related_work_table}.

A typical \mad protocol involves two or more agents to explore a diverse set of solutions, provide evidence to support their respective solutions, and reach a consensus~\citep{du2023improving}.
A judge agent can also be incorporated to read the transcripts of the debate and to give a final answer~\citep{khan2024debating}. \mad has demonstrated great potential in improving the reasoning and reducing the hallucinations of LLMs~\citep{du2023improving,liang2023encouraging,yin-etal-2023-exchange,chen-etal-2024-comm}.
\citet{liang2023encouraging} further assigned personalities to the debating agents to explore the potential answers more efficiently.
\citet{yin-etal-2023-exchange} manually specified diverse roles to agents, and proposed a confidence-based mechanism to reduce the error propagation during reasoning.
\citet{chen-etal-2024-comm} proposed a more fine-grained role assignment based on reasoning paths of agents.

Meanwhile, \mad also demonstrated great potential in providing supervision and error signals to superhuman LLMs~\citep{ai_safety_debate,khan2024debating,kenton2024on}. \citet{khan2024debating} showed that LLMs tend to be more persuasive when debating for the correct answer. \citet{kenton2024on} further showed that \mad can enable scalable oversight where a weak agent can easily tell the correctness of strong agents during debating. The success and huge potential of \mad in error detection also motivates us to use \real~\citep{realmistake} as the primary testbed to study the usefulness of \mad. Our taxonomy also has a counterpart in the study of human argumentation~\citet{walton1995commitment}: \copmad resembles a persuasion dialogue, where each party aims to convince the other of its own thesis, while the goal structure of \ours resembles an inquiry dialogue, where the participants share the aim of establishing whether a proposition is true. Different from \citet{walton1995commitment}, we study debates among LLM agents, identify the failures of existing MAD protocols through their incentive structures, and derive a proper protocol design.

\textbf{Pitfalls of Multi-Agent Debate.} Recent studies also showed the limitations of \mad~\citep{wang-etal-2024-rethinking-bounds,Smit2023ShouldWB,Zhang2025IfMD}. \citet{wang-etal-2024-rethinking-bounds} found that  \sa methods can already perform competitively or outperform \mad when given sufficient information. \citet{Smit2023ShouldWB} showed that \mad can underperform  \sa when scaling \sa using methods such as self-consistency~\citep{wang2023selfconsistency}. \citet{Zhang2025IfMD} and \citet{Yang2025RevisitingMD,kim2026capable} provided more evidence to support the findings by~\citet{wang-etal-2024-rethinking-bounds} and \citet{Smit2023ShouldWB}. These results render the effectiveness of \mad more elusive under the same budgets when compared to \sa methods.

Different from previous \mad studies, in this work, we focus on re-evaluating the usefulness of \mad through the task of error detection, which provides us new understandings of when and why \mad succeeds and fails.
Similar to~\citet{wang-etal-2024-rethinking-bounds},~\citet{Smit2023ShouldWB},~\citet{Zhang2025IfMD}, and~\citet{Yang2025RevisitingMD}, we find that the vanilla \mad protocol can often lead to degraded performances, i.e., lower than single-agent approaches. Our work goes beyond the empirical observations with a theoretical formulation of \mad failures, which further provides guidance on how to design a proper \mad protocol.

\section{Experiments}
\label{sec:exp}
We validate the theoretical predictions from Sec.~\ref{sec:mad_theory} and demonstrate the effectiveness of \ours.

\subsection{Experimental Setup}
\textbf{Datasets.} We mainly use the \real benchmark~\citep{realmistake}, which focuses on objective LLM error detection of three tasks:
(a) \mathw: The original LLM is instructed to generate a math word problem that follows the given requirements. Mistakes of LLMs can be made in following the requirements, as well as in the mathematical reasoning; (b) \fine: The original LLM is instructed to check whether the claims in a sentence are well-supported. Mistakes of LLMs can be made due to the reasoning and use of the context information; (c) \answer: The original LLM is instructed to classify whether a factual question is answerable or not. Mistakes of LLMs can be made due to hallucination and reasoning.
The original LLMs are \texttt{GPT-4-0613} and \texttt{Llama-2-70b}.
The statistics of the \real benchmark can be found in Appendix~\ref{appdx:exp}. To demonstrate generality, we additionally evaluate on GSM8K~\citep{GSM8K}, AIME-2024~\citep{AoPS_AIME}, and Anthropic-Harmful~\citep{zeng2024autodefense}, following~\citet{Yang2025RevisitingMD}.

\textbf{Baselines.} As our focus is to demonstrate the usefulness of \ours compared to \copmad and \cosmad, hence we mainly adopt the scheme in~\citet{kenton2024on} that demonstrated impressive capabilities in error detection for \copmad, and SoM~\citep{du2023improving} for \cosmad. For the implementation of SoM, we follow the practice of~\citet{Yang2025RevisitingMD}.
We also include another popular \copmad scheme, MP~\citep{liang2023encouraging}, that incorporates persona into agents. In addition, we also consider a simple \textbf{Voting} baseline, which can be considered the simplest collaborative \mad.
The reasoning-task and safety-task experiments in Table~\ref{tab:general-results} also follow the setting of~\citet{Yang2025RevisitingMD}.

\textbf{LLM Backbones.}
As the original LLMs benchmarked in \real~\citep{realmistake} are a bit outdated, we incorporate new frontier LLMs, including \texttt{GPT4o-mini}~\citep{openai2023gpt4omini}, \texttt{Llama3.1-70B}~\citep{meta2024llama3.1}, \texttt{Mistral-Small-2603}/\texttt{Mistral-7B-v0.3}~\citep{mistral}, \texttt{Qwen-2.5-72B}~\citep{qwen2.5}
as well as the frontier reasoning LLMs including \texttt{DeepSeek-R1}~\citep{r1} and \texttt{QwQ-32B}~\citep{qwq32b}.
LLM temperature is set to $0$ by default for reproducibility.

\textbf{Evaluation.} Following the prior practice, we report F1 score for \real, accuracy for reasoning tasks, attack success rate (ASR) ($\downarrow$) for safety tasks. As the nature of \real is the error detection, we additionally report and focus on the F2 score, which is an instantiation of the $F_\beta$ score~\citep{fbeta_score}:
$
F_\beta=(1+\beta^2)\cdot\text{precision}\cdot\text{recall}/(\beta^2\cdot \text{precision}+\text{recall}),
$
with setting $\beta$ as $2$. The F2 score emphasizes the recall rate, i.e., whether all errors in an LLM response can be detected.

\begin{table*}[t]
\vspace{-0.15in}
\selectfont
\centering
    \caption{Results for GPT-4 responses. The judge uses the same LLM as Debater\#1. The top two results are highlighted.}
    \label{tab:gpt4-results}
\resizebox{\textwidth}{!}{
\begin{tabular}{lll|*{6}{p{1.2cm}}|*{2}{p{1.2cm}}}
\toprule
    \multirow{2}{*}{Debater\#1} & \multirow{2}{*}{Debater\#2} & \multirow{2}{*}{Protocol}
    & \multicolumn{2}{c}{Math Problem}
    & \multicolumn{2}{c}{Fact Verification}
    & \multicolumn{2}{c}{Answerability}
    & \multicolumn{1}{c}{Avg. F1} & \multicolumn{1}{c}{Avg. F2} \\
    \cmidrule(l{2pt}r{2pt}){4-11}
     & & & {F1~($\uparrow$)}& {F2~($\uparrow$)}
     & {F1~($\uparrow$)} & {F2~($\uparrow$)}
     & {F1~($\uparrow$)} & {F2~($\uparrow$)}
     & \multicolumn{1}{c}{F1~($\uparrow$)} & \multicolumn{1}{c}{F2~($\uparrow$)} \\
\midrule
Human        & - & - & 90.00 & 84.91 & 95.45 & 95.45 & 90.48 & 87.96 & 89.44 & 91.98 \\
GPT4o-mini   & - & - & 78.70 & 89.10 & 75.76 & 82.78 & 70.10 & 74.73 & 74.85 & 82.20 \\
Llama3.1-70B & - & - & 79.26 & 89.96 & 76.85 & 85.15 & 68.13 & 69.98 & 74.75 & 81.70 \\
Mistral-Small-2603 & - & - & 76.24 & 78.05 & 55.17 & 50.00 & 44.62 & 37.08 & 58.68 & 55.04 \\
Qwen-2.5-72B & - & - & 78.82 & 77.73 & 38.18 & 28.77 & 42.37 & 32.98 & 53.12 & 46.49 \\
DeepSeek-R1  & - & - & \textbf{84.09} & 84.67 & \textbf{79.77} & 80.61 & 62.25 & 57.04 & 75.37 & 74.11 \\
\midrule
GPT4o-mini   & Llama3.1-70B & \ours & 78.38 & 90.06 & 75.12 & 85.47 & \textbf{75.36} & \textbf{83.33} & 76.29 & \textbf{86.29} \\
             &  & \copmad & 65.50 & 64.81 & 65.84 & 63.70 & 48.85 & 40.82 & 60.06 & 56.44 \\
             &  & \cosmad & 77.83 & 89.21 & 76.33 & 85.50 & \textbf{74.00} & \textbf{80.26} & 76.05 & 84.99 \\
             &  & MP & 76.00 & 82.43 & 73.56 & 74.59 & 52.78 & 46.91 & 67.45 & 67.98 \\
             &  & Voting & 78.34 & 88.91 & 75.62 & 83.33 & 68.78 & 72.22 & 74.25 & 81.49 \\
\midrule
Llama3.1-70B & GPT4o-mini & \ours & 78.03 & 89.88 & 77.42 & \textbf{88.98} & 72.91 & 79.74 & 76.12 & \textbf{86.20} \\
             &  & \copmad & 78.34 & 88.91 & 73.79 & 82.43 & 68.16 & 69.32 & 73.43 & 80.22 \\
             &  & \cosmad & 78.73 & 90.25 & 77.73 & 87.98 & 72.36 & 78.26 & 76.27 & 85.50 \\
             &  & MP & 74.03 & 75.79 & 71.35 & 75.00 & 60.13 & 55.56 & 68.50 & 68.78 \\
             &  & Voting & 78.34 & 88.91 & 75.62 & 83.33 & 68.78 & 72.22 & 74.25 & 81.49 \\
\midrule
GPT4o-mini   & Mistral-Small-2603 & \ours & 80.00 & 89.17 & 76.14 & 82.96 & 65.52 & 65.52 & 73.89 & 79.22 \\
             &  & \copmad & 52.94 & 45.34 & 46.88 & 39.16 & 24.07 & 17.62 & 41.30 & 34.04 \\
             &  & \cosmad & 77.45 & 84.95 & 75.51 & 82.04 & 59.17 & 58.14 & 70.71 & 75.04 \\
             &  & MP & 68.21 & 67.97 & 63.41 & 62.05 & 38.33 & 30.18 & 56.65 & 53.40 \\
             &  & Voting & 80.00 & 86.77 & 65.90 & 66.59 & 59.63 & 56.87 & 68.51 & 70.08 \\
\midrule
Llama3.1-70B & Mistral-Small-2603 & \ours & 79.07 & 89.29 & 79.43 & \textbf{89.44} & 66.30 & 67.87 & 74.93 & 82.20 \\
             &  & \copmad & 70.52 & 70.28 & 66.31 & 70.14 & 56.16 & 50.37 & 64.33 & 63.60 \\
             &  & \cosmad & 79.21 & 86.39 & 76.47 & 84.97 & 64.37 & 64.37 & 73.35 & 78.58 \\
             &  & MP & 49.21 & 40.05 & 61.64 & 56.11 & 53.90 & 47.26 & 54.92 & 47.81 \\
             &  & Voting & 76.62 & 83.33 & 68.97 & 69.93 & 58.06 & 54.09 & 67.88 & 69.12 \\
\midrule
GPT4o-mini   & DeepSeek-R1 & \ours & 82.76 & \textbf{90.52} & 76.77 & 83.89 & 70.79 & 71.75 & 76.77 & 82.05 \\
             &  & \copmad & 49.59 & 39.27 & 28.57 & 21.80 & 18.69 & 13.59 & 32.28 & 24.89 \\
             &  & \cosmad & 83.42 & 90.22 & 78.57 & 85.37 & 63.47 & 61.92 & 75.15 & 79.17 \\
             &  & MP & 72.41 & 72.41 & 51.95 & 48.90 & 34.15 & 27.34 & 52.84 & 49.55 \\
             &  & Voting & 81.22 & 87.34 & 79.57 & 83.90 & 65.91 & 66.36 & 75.57 & 79.20 \\
\midrule
Llama3.1-70B & DeepSeek-R1 & \ours & 81.95 & 90.13 & 78.26 & 87.66 & 73.91 & 76.40 & \textbf{78.04} & 84.73 \\
             &  & \copmad & 52.86 & 46.13 & 65.96 & 69.98 & 60.81 & 55.01 & 59.88 & 57.04 \\
             &  & \cosmad & \textbf{83.84} & \textbf{90.41} & 78.39 & 85.90 & 69.41 & 68.45 & \textbf{77.21} & 81.59 \\
             &  & MP & 31.48 & 23.04 & 47.15 & 38.36 & 42.52 & 34.79 & 40.38 & 32.06 \\
             &  & Voting & 80.81 & 87.15 & \textbf{80.85} & 85.78 & 65.48 & 64.10 & 75.71 & 79.01 \\
\bottomrule
\end{tabular}}
\vspace{-0.15in}
\end{table*}

\begin{table*}[t]
\vspace{-0.1in}
\selectfont
\centering
    \caption{F2 scores for GPT-4 responses over 10 seeds at temperature $1$. The best result is in bold and the second best is underlined. $^{*}$ marks the columns in which \ours outperforms \cosmad with $p<0.05$ (paired $t$ test), and \ours outperforms \copmad in every column ($p<10^{-4}$).}
    \label{tab:confidence-interval}
\resizebox{\textwidth}{!}{
\begin{tabular}{lll|ccc|c}
\toprule
    Debater\#1 & Debater\#2 & Protocol & Math Problem & Fact Verification & Answerability & Avg.\ F2 \\
\midrule
GPT4o-mini & - & - & $89.14\pm0.73$ & $78.71\pm2.22$ & $75.06\pm3.87$ & $80.97\pm1.53$ \\
Llama3.1-70B & - & - & $87.11\pm1.36$ & $82.31\pm1.72$ & $62.52\pm4.12$ & $77.31\pm1.43$ \\
Mistral-Small-2603 & - & - & $78.42\pm6.27$ & $61.30\pm3.46$ & $47.40\pm7.86$ & $62.37\pm4.68$ \\
\midrule
GPT4o-mini & Llama3.1-70B & \ours & $\underline{89.47\pm0.90}$ & $\mathbf{85.08\pm1.65}$ & $\mathbf{76.05\pm3.94}^{*}$ & $\mathbf{83.53\pm1.57}^{*}$ \\
 & & \copmad & $76.75\pm3.09$ & $63.78\pm4.64$ & $45.69\pm4.01$ & $62.07\pm1.54$ \\
 & & \cosmad & $\mathbf{89.64\pm0.73}$ & $\underline{83.43\pm2.16}$ & $\underline{73.35\pm4.63}$ & $\underline{82.14\pm1.88}$ \\
\midrule
Llama3.1-70B & GPT4o-mini & \ours & $\mathbf{89.23\pm1.12}$ & $\mathbf{86.52\pm1.57}^{*}$ & $\mathbf{73.48\pm2.76}$ & $\mathbf{83.08\pm1.53}^{*}$ \\
 & & \copmad & $82.44\pm1.96$ & $74.42\pm2.99$ & $56.16\pm4.46$ & $71.01\pm1.07$ \\
 & & \cosmad & $\underline{89.15\pm0.76}$ & $\underline{85.28\pm1.05}$ & $\underline{72.87\pm4.01}$ & $\underline{82.44\pm1.60}$ \\
\midrule
GPT4o-mini & Mistral-Small-2603 & \ours & $\mathbf{88.06\pm1.15}^{*}$ & $\mathbf{81.08\pm2.17}^{*}$ & $\mathbf{69.16\pm4.60}^{*}$ & $\mathbf{79.43\pm1.51}^{*}$ \\
 & & \copmad & $54.30\pm5.30$ & $47.98\pm3.93$ & $26.43\pm5.50$ & $42.90\pm2.17$ \\
 & & \cosmad & $\underline{86.65\pm2.19}$ & $\underline{79.30\pm1.86}$ & $\underline{64.62\pm5.74}$ & $\underline{76.86\pm2.67}$ \\
\midrule
Llama3.1-70B & Mistral-Small-2603 & \ours & $\mathbf{87.80\pm2.33}$ & $\mathbf{85.52\pm1.28}^{*}$ & $\mathbf{66.01\pm4.37}^{*}$ & $\mathbf{79.78\pm1.97}^{*}$ \\
 & & \copmad & $65.07\pm3.94$ & $69.45\pm3.09$ & $47.77\pm4.86$ & $60.77\pm2.27$ \\
 & & \cosmad & $\underline{86.58\pm2.58}$ & $\underline{81.90\pm2.39}$ & $\underline{63.15\pm3.85}$ & $\underline{77.21\pm1.90}$ \\
\bottomrule
\end{tabular}}
\vspace{-0.15in}
\end{table*}

\subsection{Main Results}

The results of detecting errors in GPT-4 responses are given in Table~\ref{tab:gpt4-results}, and the results for Llama-2 responses are given in Appendix~\ref{appdx:llama-2}. From the results, we have the following findings:

\textbf{Finding 1: Single agents still struggle with error detection, and existing \mad protocols do not reliably improve on them.}
Although frontier LLMs have gained lots of improvements in the past year, when incorporated in error detection, even the powerful reasoning models like \texttt{DeepSeek-R1}, still suffer from a low detection rate. Consistent with Propositions~\ref{thm:pitfal_debating} and~\ref{thm:cosmad}, \copmad often leads to performance degeneration due to its zero-sum nature. When pairing \texttt{GPT4o-mini} with \texttt{DeepSeek-R1}, \copmad drops to 24.89 Avg F2, showing severe debate hacking. Similarly, MP collapses when strong reasoning models are involved.
\cosmad improves over both single agents on the two comparable pairs, but falls below the best single agent whenever the other debater is weaker.

\textbf{Finding 2: \ours consistently outperforms all \mad baselines and improves over \sa on comparable-capability pairs.}
Under multi-seed evaluation (Table~\ref{tab:confidence-interval}), \ours outperforms \cosmad on all four pairs and \copmad by 12 to 37 points. In Table~\ref{tab:gpt4-results}, \ours outperforms \copmad, SoM, and MP on all six pairs in average F2 with zero exceptions. Against the stronger single agent, \ours improves average F2 by $2.6$ and $2.1$ points over 10 seeds on the two comparable pairs, with the gain concentrated on the task where each agent trails its partner. With a much weaker partner the gain shrinks. Results on Llama-2 responses (Appendix~\ref{appdx:llama-2}) confirm generalization: even when base performance is $\sim$90\% F2, \ours never degrades below the best single agent.

\begin{figure*}[t]
    \vspace{-0.15in}
    \centering
    \subfigure[F1 under different token costs]{
        \includegraphics[width=0.31\textwidth]{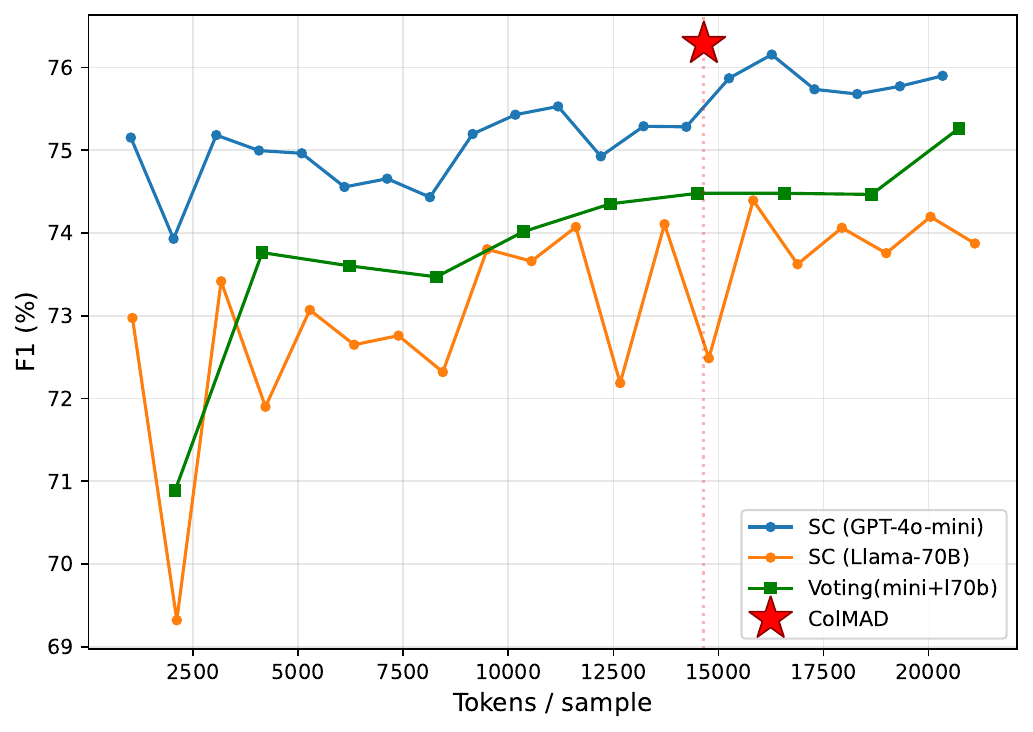}
    }
    \subfigure[F2 under different token costs]{
        \includegraphics[width=0.31\textwidth]{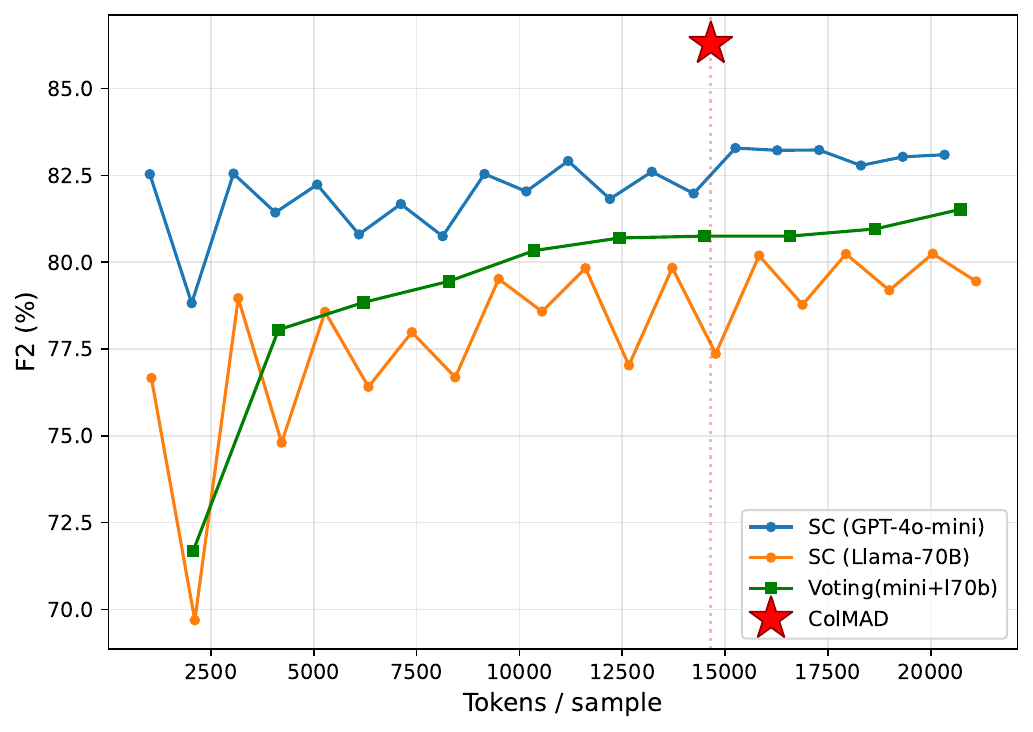}
    }
    \subfigure[F2 under different LLM calls]{
        \includegraphics[width=0.31\textwidth]{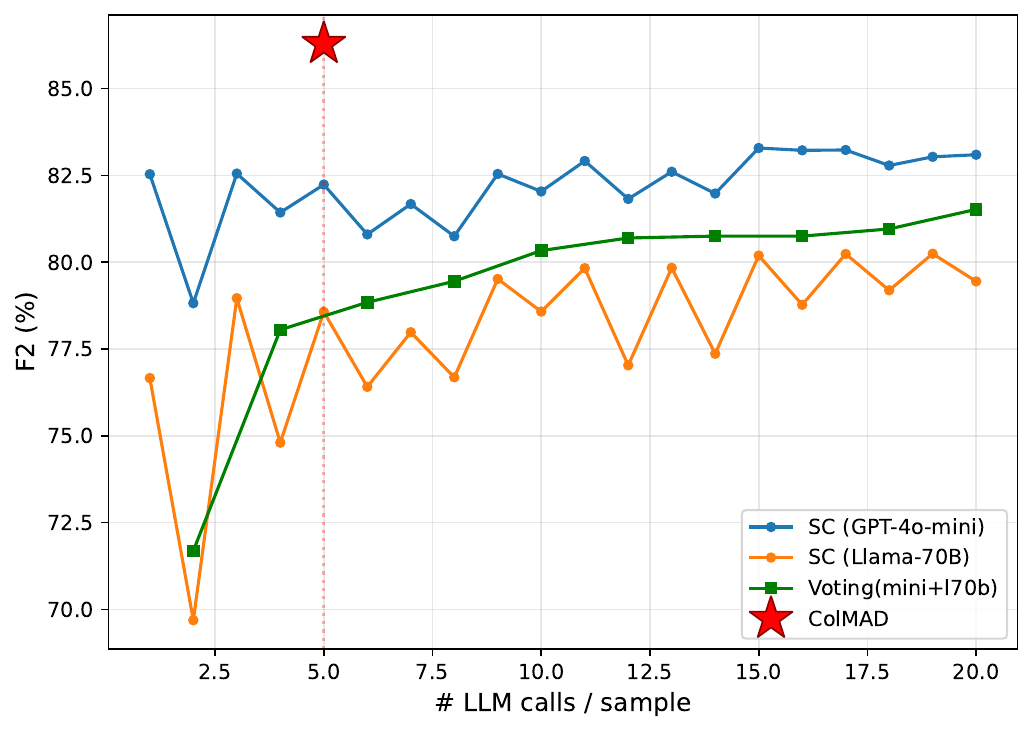}
    }
    \vspace{-0.15in}
    \label{fig:budget_control}
    \caption{Scaling \sa methods using self-consistency improves the performance a bit while saturating in the end. Under the same LLM call or token cost budget, \ours breaks the plateau of \sa, showing the potential of \mad approaches in combining complementary information from different LLMs.}
\vspace{-0.15in}
\end{figure*}
\textbf{Finding 3: \ours outperforms single-agent scaling and scaled voting under matched compute.}
A natural concern is whether \ours's gains come from using more tokens. As shown in Fig.~\ref{fig:budget_control}, we compare \ours with \sa using self-consistency (SC) and with Voting scaled by sampling both agents $m$ times. One can find that, SC saturates quickly for both models, with F2 oscillating regardless of budget, and scaled Voting saturates at about 82 average F2 by 28 calls. At
matched tokens ($\sim$14.7K), \ours outperforms SC@14 for GPT4o-mini  and Llama. Under the same LLM calls, the gap is larger still, and \ours outperforms Voting even at 28 calls. This confirms that \ours's advantage comes from effective exploitation of the \textit{cross-model diversity} instead of compute scaling.

\begin{figure*}[ht]
\vspace{-0.1in}
    \centering
    \subfigure[Gain vs.\ disagreement rate and gate]{
        \includegraphics[width=0.384\textwidth]{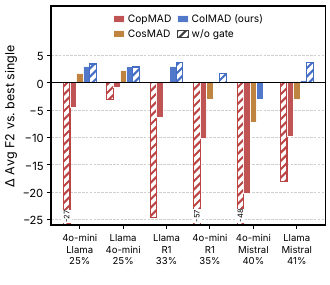}
        \label{fig:het}
    }
    \subfigure[Claim ratings]{
        \includegraphics[width=0.220\textwidth]{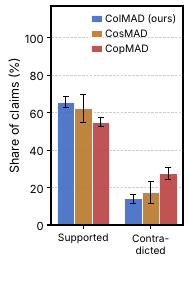}
        \label{fig:claims}
    }
    \subfigure[How \cosmad resolves disagreements]{
        \includegraphics[width=0.34\textwidth]{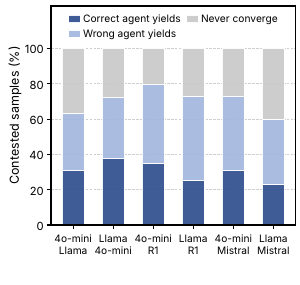}
        \label{fig:yield}
    }
    \vspace{-0.15in}
    \caption{(a)~Change in average F2 over the per-task best single agent. (b)~Share of claims in the transcripts of contested samples rated supported or contradicted by an independent rater (mean $\pm$ sd); (c)~How \cosmad resolves each contested sample of the six pairs.}
    \label{fig:analysis}
\end{figure*}
\textbf{Finding 4: \ours is the most stable protocol as disagreement grows, and elicits the most supported claims on contested samples.}
Algorithm~\ref{alg:col} skips the debate whenever the two initial answers agree, while \copmad debates every sample as originally specified~\citep{khan2024debating,kenton2024on}. Adding the gate to \copmad raises its performance, yet gated \copmad still falls below \ours on every pair, and below \ours without the gate as well.
As the initial disagreement rate grows from 25\% to 40\%, the performances of \copmad and \cosmad vary more than \ours. On the contested samples,  \ours has the largest share of supported claims and the most decisive claims per sample, i.e., supported claims that settle the verdict by themselves, while \copmad has the largest share of contradicted claims, consistent with Sec.~\ref{sec:mad_theory}. \cosmad resolves the contested samples largely independently of correctness, consistent with Proposition~\ref{thm:cosmad}.

\subsection{Ablation Studies and Analysis}
In Table~\ref{tab:general-results}, we extend to reasoning and safety benchmarks following~\citet{Yang2025RevisitingMD}, using Qwen2.5-7B and Llama3.1-8B.
\ours improves over \cosmad and the best \sa on the harder AIME-2024 at the highest call budget and dramatically reduces the attack success rate on Anthropic-Harmful to near 0\% with cross-model debate, suggesting strong potential for safety-critical oversight. On saturated tasks like GSM8K, improvements saturate, consistent with Proposition~\ref{thm:col}: little room remains for complementary information.

\begin{table*}[!t]
\vspace{-0.1in}
\selectfont
\centering
    \caption{Performance of different test-time scaling methods under different LLM call budgets.}
    \label{tab:general-results}
\resizebox{\textwidth}{!}{
\begin{tabular}{lll|*{9}{p{1.2cm}}}
\toprule
    \multirow{2}{*}{Debater\#1} & \multirow{2}{*}{Debater\#2} & \multirow{2}{*}{Protocol}
    & \multicolumn{3}{c}{GSM8K (ACC $\uparrow$)}
    & \multicolumn{3}{c}{AIME-2024 (ACC $\uparrow$)}
    & \multicolumn{3}{c}{Anthropic-Harmful (ASR $\downarrow$)} \\
    \cmidrule(l{2pt}r{2pt}){4-6}\cmidrule(l{2pt}r{2pt}){7-9}\cmidrule(l{2pt}r{2pt}){10-12}
     & & & {4} & {8} & {16} & {4} & {8} & {16} & {2} & {4} & {8} \\
\midrule
Qwen2.5-7B   & -            & -        & \textbf{90.40} & \textbf{91.40} & \textbf{90.80} & 13.33 & 13.33 & 13.33 & 14.59 & 18.92 & 31.69 \\
Llama3.1-8B  & -            & -        & 88.80 & 89.60 & 89.40 & 3.33 & 6.67 & 6.67 & 14.32 & 9.93 & 15.34 \\
Qwen2.5-7B   & Llama3.1-8B  & \ours    & 90.00 & 90.20 & 90.60 & 10.00 & 10.00 & \textbf{20.00} & \textbf{0.00} & \textbf{0.00} & \textbf{0.27} \\
Qwen2.5-7B   &  Llama3.1-8B            & \cosmad  & 88.00 & 87.80 & 87.80 & 6.67 & 6.67 & 6.67 & 20.27 & 20.54 & 23.24 \\
Qwen2.5-7B   & Llama3.1-8B             & \copmad       & 88.40     & 47.40     & 56.40     & \textbf{16.67} & 3.33 & 3.33 & 7.30 & 23.51 & 30.27 \\
\bottomrule
\end{tabular}}
\vspace{-0.15in}
\end{table*}

\begin{figure*}[ht]
    \vspace{-0.15in}
    \centering
    \subfigure[Combination of different LLMs]{
        \includegraphics[width=0.6\textwidth]{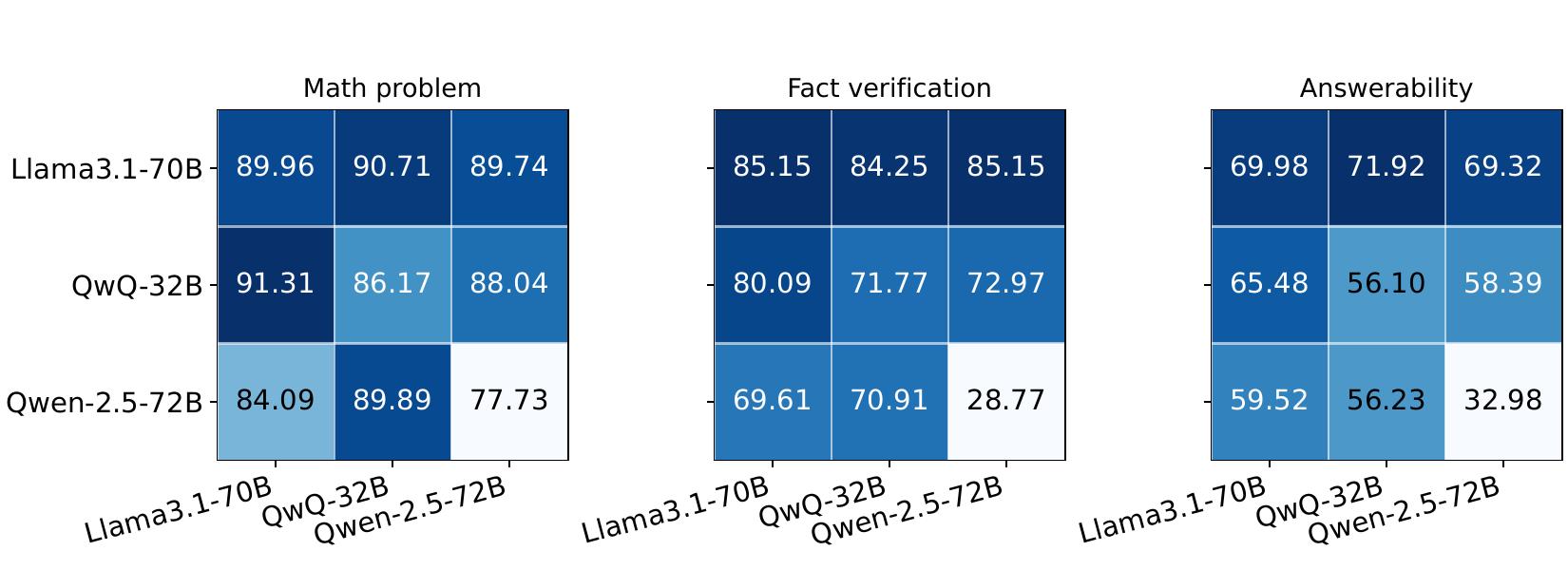}
        \label{fig:llm_comb}
    }
    \subfigure[Explanation alignment]{
        \includegraphics[width=0.35\textwidth]{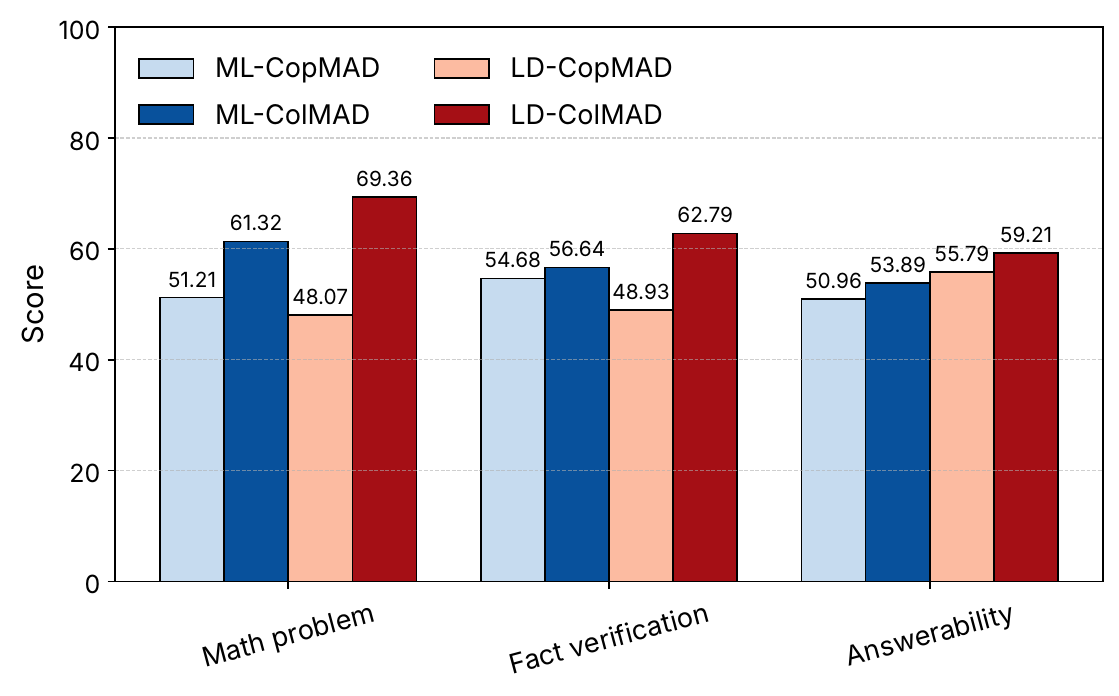}
        \label{fig:rating}
    }
    \vspace{-0.15in}
    \caption{(a) shows the results in F2 scores of \ours performance under different combinations of LLMs, where the diagonal line shows the single-agent performance. (b) shows the rate of alignment to the ground-truth explanations given by \copmad and \ours, where ``ML-" refers to the combination of \texttt{GPT4o-mini} and \texttt{Llama3.1-70B}, and ``LD-" refers to the combination of \texttt{Llama3.1-70B} and \texttt{DeepSeek-R1}. \ours yields more reasonable explanations than \copmad.}
    \label{fig:explanation}
    \vspace{-0.15in}
\end{figure*}
\begin{figure*}[!ht]
    \centering
    \vspace{-0.15in}
    \subfigure[Math problem]{
        \includegraphics[width=0.31\textwidth]{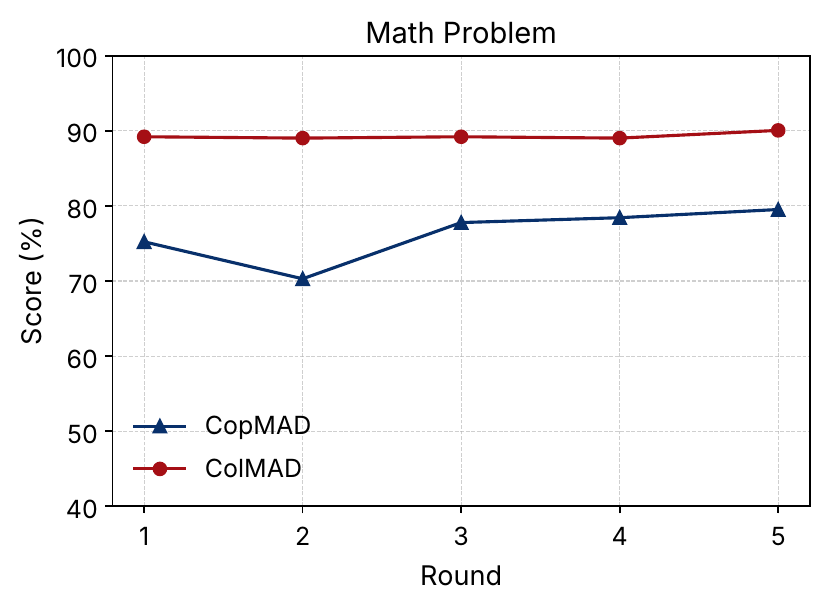}
    }
    \subfigure[Fact verification]{
        \includegraphics[width=0.31\textwidth]{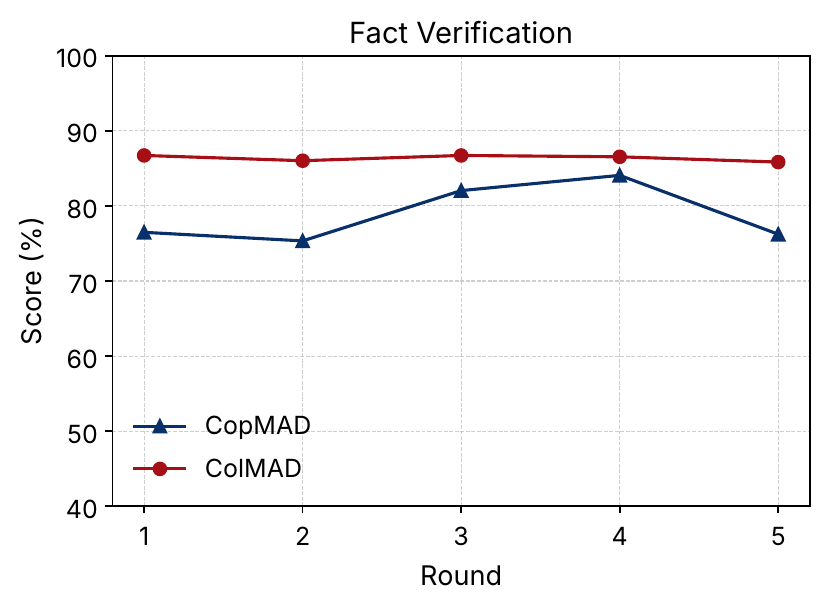}
    }
    \subfigure[Answerability]{
        \includegraphics[width=0.31\textwidth]{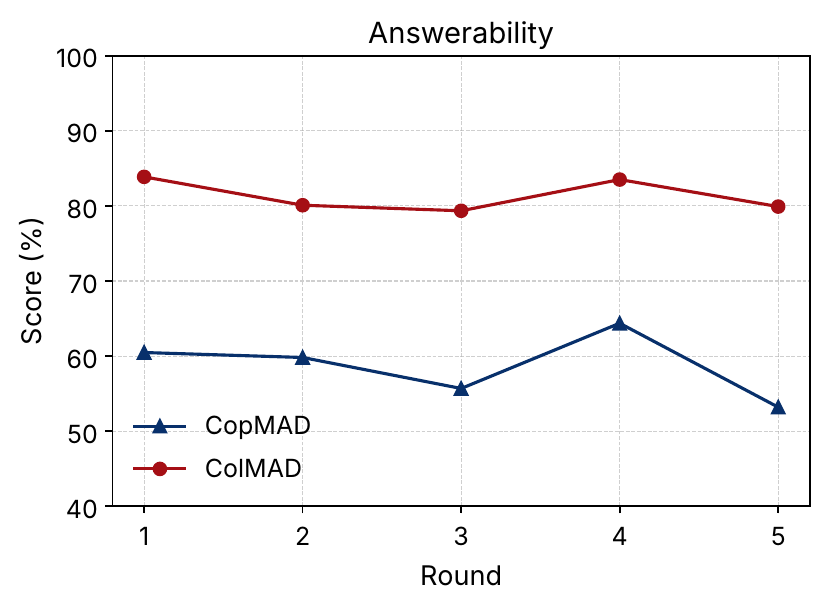}
    }
    \vspace{-0.05in}
    \caption{\ours is generically robust to the number of rounds for debate.}
    \label{fig:debating-rounds}
    \vspace{-0.15in}
\end{figure*}

\textbf{Judge robustness.} Different combinations of LLMs as judges are shown in Fig.~\ref{fig:llm_comb}. Swapping the judge changes \ours's Avg F2 by only 0.03, while \copmad varies by 22.88, indicating that \ours produces balanced transcripts that any competent judge can evaluate.

\textbf{Explanation alignment.} Fig.~\ref{fig:rating} shows \ours yields more human-aligned explanations than \copmad, critical for scalable oversight. Fine-grained alignment results are given in Appendix~\ref{appdx:finegrained}.

\textbf{Sensitivity to debate rounds.}
In Fig.~\ref{fig:debating-rounds}, \ours is robust to the number of debate rounds, while \copmad degrades with more rounds. The results are also consistent with babbling accumulation under competitive incentives, where the debate messages become more uninformative with more rounds.

\textbf{More results.} We also provide ablation studies on different prompt components in Appendix~\ref{appdx:prompt}, as well as different debater settings in Appendix~\ref{appdx:same_debater}.

\section{Conclusions}
In this work, we investigated when and why \mad fails by examining the incentive structures of \copmad and \cosmad protocols. We showed that both suffer from debate hacking due to goal misalignment: \copmad incentivizes persuasion over truth-seeking, while \cosmad suppresses informative disagreements. We introduced \ours, a new collaborative \mad protocol that encourages agents to provide informative and truthful messages.
Our results on error detection, together with extended evaluations on reasoning and safety tasks, show that \mad is not inherently ineffective: with heterogeneous debaters and a proper protocol design, \mad can improve over \sa methods.

\clearpage
\begin{ack}
We thank the reviewers and the area chair for their constructive comments, which directly shaped the additional analyses in this paper.
MS was supported by JST ASPIRE Grant Number JPMJAP2405.
BH was supported by NSFC General Program No. 62376235, RGC General Research Fund No. 12200725, RGC Research Fellow Scheme No. RFS2627-2S02, RGC Young Collaborative Research Grant No. C2005-24Y, RGC General Research Fund No. 12202026, NSFC Major Research Plan No. 92570109, RIKEN Collaborative Research Fund, Guangdong and Hong Kong Universities ``1+1+1'' Joint Research Collaboration Scheme, HKBU Faculty Niche Research Areas No. RC-FNRA-IG/25-26/SCI/05, and HKBU CSD Departmental Incentive Scheme.
\end{ack}

\bibliography{references/mad,references/llm,references/graphllm,references/mllm,references/graphood,references/ood,references/references,references/error_detection,references/w2s}
\bibliographystyle{abbrvnat}

\clearpage
\onecolumn
\appendix
\section*{LLM Use Statement}
From the research side, this work studies the use of LLMs to perform multi-agent debate to detect errors in LLM responses. From the paper writing side, we use LLMs to assist with improving the writing of this work.

\section*{Broader impacts}
This work does not involve human subjects or personally identifiable information beyond public benchmarks used under their licenses. Our experiments evaluate error detection and decision protocols among LLMs, which benefits the oversight and prevents potential risks of superhuman intelligence in the future.

\section{Discussion and Future Work}
\label{appdx:discussion}

\paragraph{Training for collaborative debate.}
The current implementation relies on prompting to elicit collaborative behavior, but LLMs are predominantly trained on single-agent objectives. Fine-tuning debater models with multi-agent objectives, e.g., rewarding information gain $I(Y; m_i | X_0, M^{(t-1)})$ rather than persuasion probability could internalize the collaborative incentive structure.

\paragraph{Honesty and uncertainty calibration.}
While \ours demonstrates that collaborative debate can overcome the limitations of competitive and consensus-seeking \mad protocols, our behavioral analysis also reveals its remaining failure modes, e.g., there are still some overconfident claims. This reflects a fundamental calibration problem: LLMs express high confidence on claims they cannot verify. Training for honesty through objectives that penalize confident assertions on uncertain claims, would reduce both overconfidence and fabricated evidence simultaneously.

\paragraph{Broader applications.}
We evaluate \ours on error detection, mathematical reasoning, and safety alignment, but the collaborative debate framework is applicable to any task where cross-checking between agents can surface complementary information, including code review, scientific claim verification, and safety-critical content moderation. Extending to multi-party debate (beyond two agents) and heterogeneous agent roles (e.g., specialist verifiers for different requirement types) are natural next steps.

\section{More Details about Theories}
\label{appdx:theory}

\subsection{Notations}
A table of notations used in our work is given as follows.

\begin{table}[ht]\label{tab:notation}
\centering
\small
\caption{Table of Notations.}
\resizebox{\textwidth}{!}{
\begin{tabular}{ll}
\toprule
\textbf{Notation} & \textbf{Meaning} \\
\midrule
$y\in\{0,1\}$ & True label (e.g., \textit{error} vs \textit{no\_error}). \\
$Y$ & Random variable of the label predictions. \\
$\vy$ & Predictions of the labels over a dataset. \\
$\pi$ & Prior $\Pr(y=1)$; prior log-odds $\log\frac{\pi}{1-\pi}$. \\
$X_0=(a,b)$ & Baseline signals from the two base models A, B (what the judge has without debate). \\
$p(x\mid y)$ & Likelihood of baseline signal $x$ under label $y$. \\
$\Lambda_0(x)=\log\frac{p(x\mid y=1)}{p(x\mid y=0)}$ & Baseline log–likelihood ratio (LLR), i.e., weight of evidence from $X_0$. \\
$m_A,\,m_B$ & Debate messages emitted by debater A and B. \\
$M=(m_A,m_B)$ & Joint debate messages. \\
$p_i(m_i\mid y,x)$ & Conditional likelihood model of debater $i$'s message given $y$ and $x=X_0$. \\
$\ell_i(m_i;x)=\log\frac{p_i(m_i\mid y=1,x)}{p_i(m_i\mid y=0,x)}$ & Debater $i$'s additive LLR contribution given message and context. \\
$|\ell_i|\le L_i$ & Bounded manipulability / persuasion budget for debater $i$. \\
$\Lambda(x,m_A,m_B)$ & Total LLR used by the judge: \\
& $\displaystyle \Lambda_0(x)+\ell_A(m_A;x)+\ell_B(m_B;x)+\log\frac{\pi}{1-\pi}$. \\
$J$ & Judge decision rule mapping $(X_0,m_A,m_B)\mapsto\{0,1\}$. \\
$R(Z)$ & Bayes (minimum) $0$–$1$ risk achievable when using signal $Z$. \\
$R_\mathrm{base}:=R(X_0)$ & Baseline Bayes risk using only the base signals (no debate). \\
$V_\mathrm{adv}$ & Minimax error in adversarial (zero-sum) debating. \\
$R_\mathrm{coop}$ & Bayes risk under cooperative, truth-seeking debating (messages add true evidence). \\
$I(y;M\mid X_0)$ & Conditional mutual information: new information from messages beyond $X_0$. \\
$\eta(z)=\Pr(y=1\mid z)$ & Posterior probability under observable $z$ (e.g., $z=X_0$ or $z=(X_0,M)$). \\
$\frac{1}{1+e^{|\Lambda|}}$ & Instantaneous Bayes error at balanced prior ($\pi=\tfrac12$) for total LLR $\Lambda$. \\
\bottomrule
\end{tabular}}
\end{table}

\paragraph{Debate setup.}
We consider two agents $\mathrm{A}$ (Alice) and $\mathrm{B}$ (Bob), and denote the predictions of the error detections as $\hat{\vy}_A$ and $\hat{\vy}_B$, with rationales (e.g., CoT reasoning) as $x_A$ and $x_B$, respectively. During the debate, they will emit messages $m_A$ and $m_B$ to convince a judge $J$.
\begin{assumption}[Optimal Judge Strategy]\label{assump:optimal_judge_appdx}
Denoting the label as $Y\in\{0,1\}$ with prior probability $\pi\in(0,1)$, without debating, the judge will derive the final answer based on the initial responses $X_0=(x_A,x_B)$. Assuming the judge $J$ uses the Bayes test on the total log-likelihood ratio (LLR), we will have the LLR based on $X_0$ as
\begin{equation}\label{eq-appdx:llr_x0}
    \Lambda_0(X_0)=\log\frac{p(X_0|Y=1)}{P(X_0|Y=0)}+\log\frac{\pi}{1-\pi}.
\end{equation}
The contribution to LLR from the debating can be written as
\begin{equation}\label{eq-appdx:llr_contrib}
    l_i(m_i;X_0)=\log\frac{P(m_i|Y=1,X_0)}{P(m_i|Y=0,X_0)},\ i\in\{A,B\}.
\end{equation}
Then, with debating, the LLR of the judge is
\begin{equation}\label{eq-appdx:llr_debating}
\Lambda(X_0,m_A,m_B)=\Lambda_0(X_0)+l_A(m_A;X_0)+l_B(m_B;X_0)+\log\frac{\pi}{1-\pi}.
\end{equation}
\end{assumption}

Intuitively, if the elicited messages $M=(m_A,m_B)$ bring additional information to the judge, i.e., $I(Y;M|X_0)>0$ where $I(\cdot;\cdot)$ denotes the mutual information, we will have $\Lambda(X_0,M)>\Lambda(X_0)$.
In order to provide additional information, we need to look into the cases where the predictions by $\mathrm{A}$ differ from those by $\mathrm{B}$. Under $Y_A\neq Y_B$, if both agents are able to provide \textit{sufficient justifications} and are more persuasive during the debate when they are debating for the \textit{correct} answer. Hence, the judge can be convinced to take the correct answer when either of the agents is correct.
The reduction of errors from the oracle collaboration can be calculated through
\begin{equation}\label{eq-appdx:oracle_collab}
    \min_{K\in\{A,B\}}\sum_{i}\mathbbm{1}[\hat{\vy}_K^{(i)}\neq\vy^{(i)}]-\sum_{i}\mathbbm{1}[\hat{\vy}_J^{(i)}\neq\vy^{(i)}],
\end{equation}
where $\vy^{(i)}_K$ denotes the initial prediction of the agent $K\in\{A,B\}$ on the $i$-th sample. Intuitively, Eq.~(\ref{eq-appdx:oracle_collab}) can be considered as the potential of the collaboration between $\mathrm{A}$ and $\mathrm{B}$.

\subsection{Game-theoretic formulation of existing \mad}

Essentially, existing \mad methods can be categorized into two categories: \copmad where LLM agents competitively debate with each other to convince the judge of the respective assigned answers~\citep{khan2024debating}; and \cosmad (i.e., Consensus-seeking \mad) where LLM agents seek consensus~\citep{du2023improving}.

Then, we provide game-theoretic definitions of \copmad, \ours, and \cosmad:

\begin{definition}[Basic \mad setup]
    The basic setting of \mad games is that, given a question $Q$ with a binary label $Y\in\{0,1\}$, two debaters $\mathrm{A}$ and $\mathrm{B}$ with different beliefs of the label that debate with each other and produce messages $M=(m_A,m_B)$. Without loss of generality, $\mathrm{A}$ believes the answer $Y_A=1$ with initial rationale $x_A$ and $\mathrm{B}$ believes $Y_B=0$ with initial rationale $x_B$. A judge $J$ will give Bayes-optimal predictions $Y_J=J(X_0,M)$ based on the information $X_0$ and $M$ (Assumption~\ref{assump:optimal_judge_appdx}).
    When we consider the number of debating rounds, we will denote the messages generated before round $t$ as $M^{(t-1)}$; otherwise, we will omit it for the simplicity of notations.
\end{definition}
\begin{definition}[Competitive \mad]
	Debaters $\mathrm{A}$ and $\mathrm{B}$ have opposed utilities:
	$u_A(m_A)=P(Y_J=Y_A\mid m_A,X_0,M^{(t-1)}),\  u_B(m_B)=P(Y_J=Y_B\mid m_B,X_0,M^{(t-1)}).$
	The judge observes the full transcript $M$ and the risk is $V_\mathrm{cop} = \min_J \max_{e \in \mathcal{E}_\mathrm{cop}(J)} P(J(X_0, M_e) \neq Y)$, where $M_e$ is the debating transcripts given by the optimal Nash equilibrium $e \in \mathcal{E}_\mathrm{cop}(J)$ of the A-B subgame in convincing $J$.
\end{definition}

\begin{definition}[Consensus-seeking \mad]
Debaters $\mathrm{A}$ and $\mathrm{B}$ share the same utilities towards reaching consensus:
$u_A(m_A) = u_B(m_B) = P(Y_A = Y_B \mid
X_0, M^{(t-1)})$.
The risk is
$V_\mathrm{cos} = \min_J \min_{e \in
\mathcal{E}_\mathrm{cos}(J)}
P(J(X_0, M_e) \neq Y)$, where $M_e=\Phi_\cap(M)$ is the debating transcripts given by the optimal Nash equilibrium $e \in \mathcal{E}_\mathrm{cos}(J)$ of the A-B subgame in reaching consensus.
\end{definition}

\begin{definition}[Collaborative \mad]
Debaters have truth-seeking utilities: $u_A(m_A)=u_B(m_B)=I(Y;m_i\mid X_0,M^{(t-1)})$. The judge observes the full transcript $M$. The risk is $V_\mathrm{col}=\min_J\min_{e\in\mathcal{E}_\mathrm{col}(J)}P(J(X_0,M_e)\neq Y)$.
\end{definition}

\subsection{Proof for Proposition~\ref{thm:pitfal_debating} (Competitive \mad)}
\label{proof:pitfal_debating}
\begin{proposition}[Restatement of Proposition~\ref{thm:pitfal_debating}]%
    Assuming a bounded LLR, i.e., $l_i\leq L_i<\infty,\ i\in\{A,B\}$, let $R(Z)$ denote the Bayes risk of the judge $J$ when making decisions based on $Z$, denote $R_0=R(X_0)$, denote the outcome of zero-sum debating as
    \[
    V_\mathrm{cop}=\min_J\max_{e\in\mathcal{E}_\mathrm{cop}(J)} P(J(X_0,M_e)\neq Y),
    \]
    where $M_e$ are the debating transcripts given by the optimal Nash equilibrium $e\in\mathcal{E}_\mathrm{cop}(J)$ of the $\mathrm{A}$–$\mathrm{B}$ subgame in convincing $J$.
    Then, we have $V_\mathrm{cop}=R_0$.
\end{proposition}
\begin{proof}
    To show $V_\mathrm{cop}=R_0$, we need to show $V_\mathrm{cop}\leq R_0$, and $V_\mathrm{cop}\geq R_0$, and provide a condition that the equity holds.

    (i) For $V_\mathrm{cop}\leq R_0$, given that the judge aims to choose the optimal strategy given any strategies of $\mathrm{A}$ and $\mathrm{B}$ that may degenerate the debate performance. We first consider a simple ignore strategy for the judge, $J_\mathrm{ignore}$ that directly drops any additional debate transcripts between $\mathrm{A}$ and $\mathrm{B}$. We have
    \begin{equation}
        P(J_\mathrm{ignore}(X_0,M_e)\neq Y)=P(J_\mathrm{ignore}(X_0)\neq Y)=R_0,
    \end{equation}
    which follows
    \begin{equation}
        \max_{e\in\mathcal{E}_\mathrm{cop}(J)}P(J_\mathrm{ignore}(X_0,M_e)\neq Y)=R_0.
    \end{equation}
    Then, it suffices to know that
    \begin{equation}
        V_\mathrm{cop}=\min_J\max_{e\in\mathcal{E}_\mathrm{cop}(J)} P(J(X_0,M_e)\neq Y)\leq \max_{e\in\mathcal{E}_\mathrm{cop}(J)}P(J_\mathrm{ignore}(X_0,M_e)\neq Y)=R_0,
    \end{equation}
    and $V_\mathrm{cop}\leq R_0$.

    (ii) For $V_\mathrm{cop}\geq R_0$, we consider the strategies of the debaters. Without loss of generality, we assume $\mathrm{A}$ is debating for $Y=1$ and $\mathrm{B}$ is debating for $Y=0$. Since we do not impose any limits on the capabilities of the debater agents, they will try to present the evidence as most useful for the respective answer as they can. More formally, the optimal strategies for debaters $\mathrm{A}$ and $\mathrm{B}$ will yield the following
    \begin{equation}
        M\mathrel{\perp\!\!\!\perp} Y |X_0.
    \end{equation}
    It follows that
    \begin{equation}
        P(Y=1|X_0,M)=P(Y=1|X_0).
    \end{equation}
    Therefore, it suffices to know that $V_\mathrm{cop}\geq R_0$. That concludes our proof.
\end{proof}

\paragraph{Connection to cheap-talk games.} The structure of \copmad --- costless messages between misaligned senders and a receiver --- is analogous to the cheap-talk game of~\citet{crawford1982strategic}. The equilibrium $M\perp Y|X_0$ corresponds to the \textit{babbling equilibrium} in cheap-talk theory, where messages carry no information because the sender's incentive is to persuade rather than inform. This provides a game-theoretic explanation for the debate hacking phenomenon observed in Fig.~\ref{fig:pitfall_mad}.

\subsection{Proof for Proposition~\ref{thm:cosmad} (Consensus-Seeking \mad)}
\label{proof:cosmad}

\begin{proposition}[Pitfalls of consensus-seeking debate (restated)]
Let $R(X_0, Z)$ denote the Bayes risk when the judge decides based on $(X_0, Z)$. Then $R(X_0, \Phi_\cap(M)) \geq R(X_0, M)$, with strict inequality when $I(Y;M|X_0) > I(Y;\Phi_\cap(M)|X_0)$.
\end{proposition}

\begin{proof}
Since $\Phi_\cap(M)$ is a deterministic function of $M$ (retaining only the subset on which both agents agree), by the data processing inequality:
\begin{equation}
    I(Y;\Phi_\cap(M)|X_0)\leq I(Y;M|X_0).
\end{equation}
The Bayes risk $R(X_0, Z)$ is a monotonically decreasing function of $I(Y;Z|X_0)$: more information about $Y$ allows a better decision. Therefore:
\begin{equation}
    R(X_0, \Phi_\cap(M))\geq R(X_0, M).
\end{equation}
The strict inequality holds when $\Phi_\cap$ discards information about $Y$, i.e., $I(Y;M|X_0)>I(Y;\Phi_\cap(M)|X_0)$. This occurs when there exist disagreements between agents that carry information about the true label but are filtered out by the consensus process.
\end{proof}

\paragraph{Interpretation.} The information loss equals the conditional mutual information between $Y$ and the disagreements, given the consensus:
\begin{equation}
    I(Y;M|X_0) - I(Y;\Phi_\cap(M)|X_0) = I(Y;M\setminus\Phi_\cap(M) \mid X_0, \Phi_\cap(M)).
\end{equation}
This quantity is strictly positive whenever the disagreements carry information about $Y$ that is not already captured by the consensus, which formalizes the intuition that \textit{disagreements are the most valuable signals} in a debate.

\subsection{Proof for Proposition~\ref{thm:col} (Collaborative \mad)}
\label{proof:col}

\begin{proposition}[Collaborative debate (restated)]
$V_\mathrm{col}\leq R(X_0)=V_\mathrm{cop}$, with strict inequality when $I(Y;M_e|X_0)>0$.
\end{proposition}

\begin{proof}
\textbf{$V_\mathrm{col}\leq R(X_0)$:} The ignore strategy $J_\mathrm{ignore}$ gives $R(X_0)$ regardless of $M$, so $V_\mathrm{col}\leq R(X_0)$.

\textbf{Strict inequality when $I(Y;M_e|X_0)>0$:} When messages carry positive information, the posterior $\eta(X_0,M_e)=P(Y=1|X_0,M_e)$ differs from $\eta(X_0)=P(Y=1|X_0)$ with positive probability. Under 0--1 loss, $R(Z)=1-\max\{\eta(Z),1-\eta(Z)\}$. With positive probability there exists $(X_0,M_e)$ such that:
\begin{equation}
    \max\{\eta(X_0,M_e),1-\eta(X_0,M_e)\}>\max\{\eta(X_0),1-\eta(X_0)\},
\end{equation}
yielding $R(X_0,M)<R(X_0)$, so $V_\mathrm{col}<R(X_0)=V_\mathrm{cop}$.
\end{proof}

\subsection{Proof for Corollary~\ref{cor:ordering} (Protocol Ordering)}

\begin{corollary}[Ordering of \mad protocols (restated)]
$V_\mathrm{col}\leq R(X_0,\Phi_\cap(M))\leq R(X_0)=V_\mathrm{cop}.$
\end{corollary}

\begin{proof}
We combine the three propositions:
\begin{itemize}[leftmargin=1.5em]
    \item $V_\mathrm{cop}=R(X_0)$: from Proposition~\ref{thm:pitfal_debating}.
    \item $R(X_0,\Phi_\cap(M))\leq R(X_0)$: the consensus subset $\Phi_\cap(M)$ carries at least as much information as $X_0$ alone (agents may add useful information even after filtering), so $R(X_0,\Phi_\cap(M))\leq R(X_0)$.
    \item $V_\mathrm{col}\leq R(X_0,\Phi_\cap(M))$: from Proposition~\ref{thm:cosmad}, $R(X_0,\Phi_\cap(M))\geq R(X_0,M)$, and $V_\mathrm{col}=\min_J R(X_0,M)\leq R(X_0,M)\leq R(X_0,\Phi_\cap(M))$.
\end{itemize}
Therefore $V_\mathrm{col}\leq R(X_0,\Phi_\cap(M))\leq R(X_0)=V_\mathrm{cop}$.
\end{proof}

\paragraph{Interpretation.} The ordering reflects three levels of information utilization:
\begin{itemize}[leftmargin=1.5em]
    \item \copmad: opposed incentives destroy information ($I(Y;M|X_0)=0$), so the judge effectively has only $X_0$.
    \item \cosmad: agents produce some useful information, but consensus filtering discards the disagreements, leaving $\Phi_\cap(M)\subseteq M$.
    \item \ours: truth-seeking incentives produce informative messages, and the judge sees the full transcript $M$ including disagreements.
\end{itemize}
Each step from \copmad to \cosmad to \ours preserves strictly more information about $Y$, yielding strictly lower risk when the additional information is non-trivial.

\subsection{Theory-to-implementation mapping}
\label{appdx:theory_impl_map}

Table~\ref{tab:theory-impl-map} maps the theoretical condition that \ours requires for its advantage over \copmad (Proposition~\ref{thm:col}) to the specific prompt components used in our implementation.

\begin{table}[t]
\centering
\caption{Mapping between the theoretical condition for \ours (Proposition~\ref{thm:col}) and the practical prompt components in our implementation.}
\label{tab:theory-impl-map}
\small
\setlength{\tabcolsep}{6pt}
\renewcommand{\arraystretch}{1.2}
\begin{tabular}{p{0.30\linewidth} p{0.25\linewidth} p{0.38\linewidth}}
\toprule
Theoretical requirement & Practical design & Mechanism \\
\midrule
Messages carry new information about $Y$ beyond $X_0$
    & ``Complement missing points'' instruction
    & Maximizes $I(Y; M_e \mid X_0)$ by directing agents to identify what the other missed. \\
Messages are truthful, not fabricated
    & Evidence verification (quote-based)
    & Ensures debate messages reflect genuine evidence grounded in the context. \\
Reduce overconfidence and dishonesty
    & Self-auditing
    & Agents identify plausible failure modes in their own claims, approximating the honest-debater assumption. \\
Judge can properly weight evidence
    & Confidence calibration
    & Helps the judge estimate the LLR contribution of each message. \\
\bottomrule
\end{tabular}
\end{table}

\section{Categorization of existing \mad approaches}
\label{appdx:related_work_table}

We summarize the existing \mad literature in Table~\ref{tab:related-work}, covering the two dominant protocols (\copmad and \cosmad) as well as orthogonal directions (communication, application, and benchmarking). Our \ours is the first \mad protocol that reframes debate as a non-zero sum collaborative game while retaining the adversarial role of debate to surface decisive errors.

\begin{table}[t]
\centering
\caption{Categorization of existing \mad approaches. We position \ours as a collaborative \mad, distinct from the two dominant lines in the literature: \copmad (competitive debate) and \cosmad (consensus-seeking debate). Works without a \texttt{natbib} key are cited inline.}
\label{tab:related-work}
\small
\setlength{\tabcolsep}{4pt}
\begin{tabular}{lll}
\toprule
Name & \mad aspect & Reference \\
\midrule
\ours                     & Collaborative \mad                    & Ours \\
\copmad                   & Competitive \mad                      & \citet{kenton2024on} \\
\citet{khan2024debating}         & Competitive \mad                      & \citet{khan2024debating} \\
\citet{irving2018ai}       & Competitive \mad                      & \citet{irving2018ai} \\
\citet{BrownCohen2025AvoidingOW}  & Competitive \mad                      & \citet{BrownCohen2025AvoidingOW}  \\
\citet{Buhl2025AnAS}         & Competitive \mad                      & \citet{Buhl2025AnAS} \\
\citet{BrownCohen2023ScalableAS} & Competitive \mad                      & \citet{BrownCohen2023ScalableAS} \\
\citet{rahman2026ai}      & Competitive \mad                      & \citet{rahman2026ai} \\
\citet{carro2025ai}        & Competitive \mad                      & \citet{carro2025ai}  \\
MP                        & Competitive \mad with persona         & \citet{liang2023encouraging} \\
SoM                       & Consensus-seeking \mad (\cosmad)      & \citet{du2023improving} \\
Reconcile                 & Consensus-seeking \mad (\cosmad)      & \citet{chen-etal-2024-reconcile} \\
CoMM                      & Consensus-seeking \mad with persona   & \citet{chen-etal-2024-comm} \\
EoT                       & Communication protocol for \mad       & \citet{yin-etal-2023-exchange} \\
ChatEval                  & Application of \mad (evaluation)      & \citet{chan2023chateval} \\
\citet{wang-etal-2024-rethinking-bounds}         & Benchmarking of \mad                  & \citet{wang-etal-2024-rethinking-bounds} \\
\citet{Smit2023ShouldWB}         & Benchmarking of \mad                  & \citet{Smit2023ShouldWB} \\
\citet{Wynn2025TalkIA}         & Benchmarking of \cosmad               & \citet{Wynn2025TalkIA} \\
\citet{Zhang2025IfMD}        & Benchmarking of \mad                  & \citet{Zhang2025IfMD} \\
\citet{Yang2025RevisitingMD}         & Benchmarking of \mad                  & \citet{Yang2025RevisitingMD} \\
\bottomrule
\end{tabular}%
\end{table}

\section{More Results on Potential of Multi-Agent Collaboration}\label{appdx:potential}

Given \real, we provide more results on the error reduction of multi-agent collaboration under the oracle protocol as Eq.~(\ref{eq-appdx:oracle_collab}).

\begin{figure*}[ht]
    \centering
    \subfigure[Math problem]{
        \includegraphics[width=0.31\textwidth]{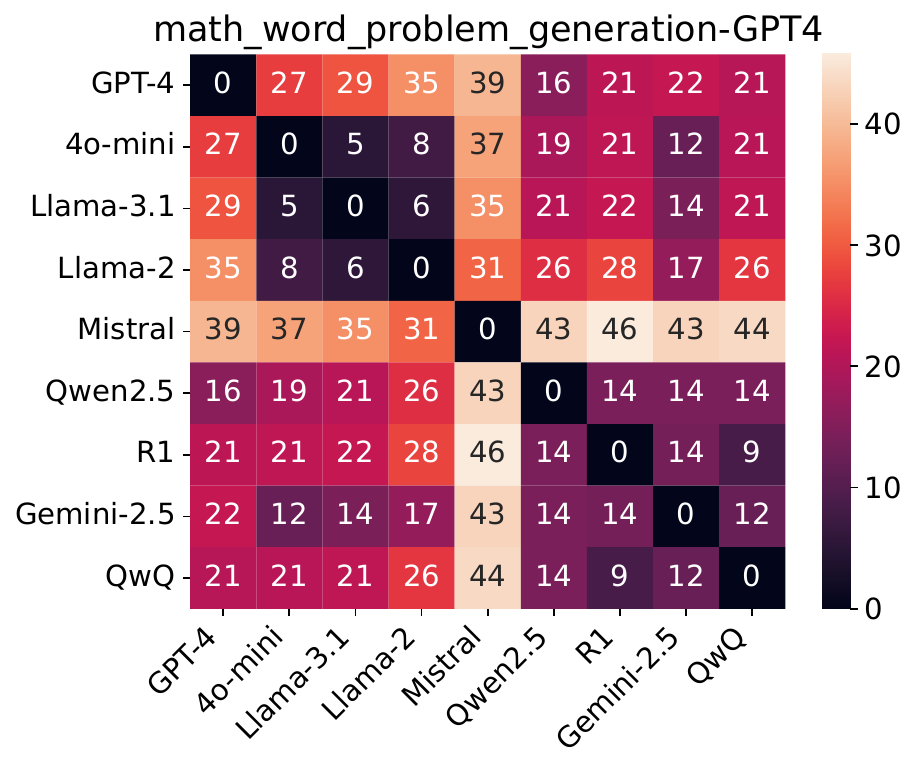}
    }
    \subfigure[Fact verification]{
        \includegraphics[width=0.31\textwidth]{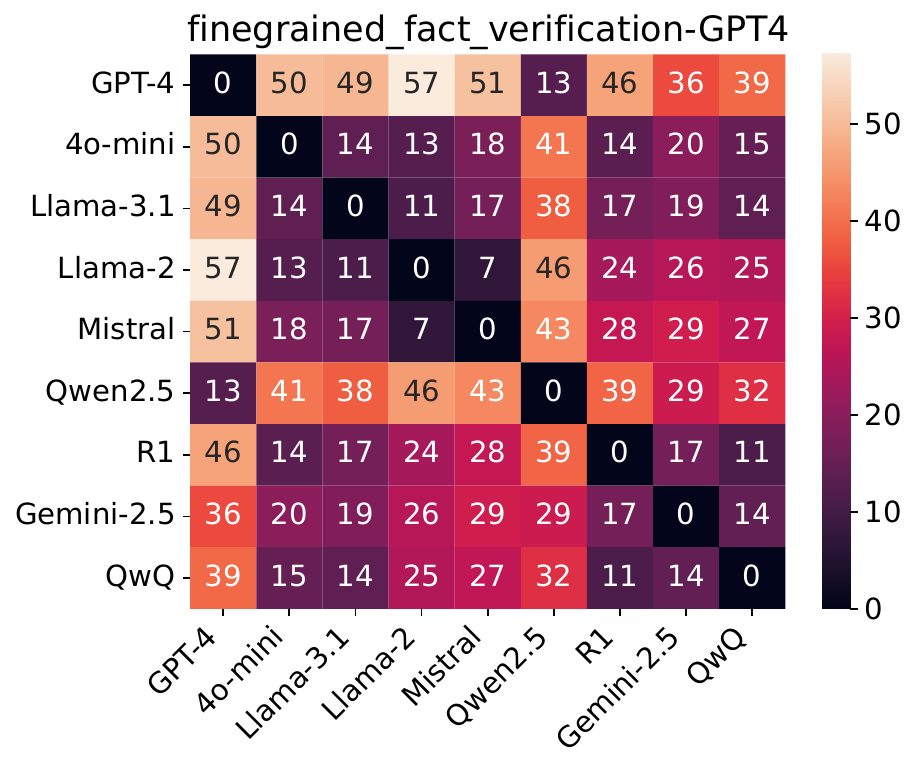}
    }
    \subfigure[Answerability]{
        \includegraphics[width=0.31\textwidth]{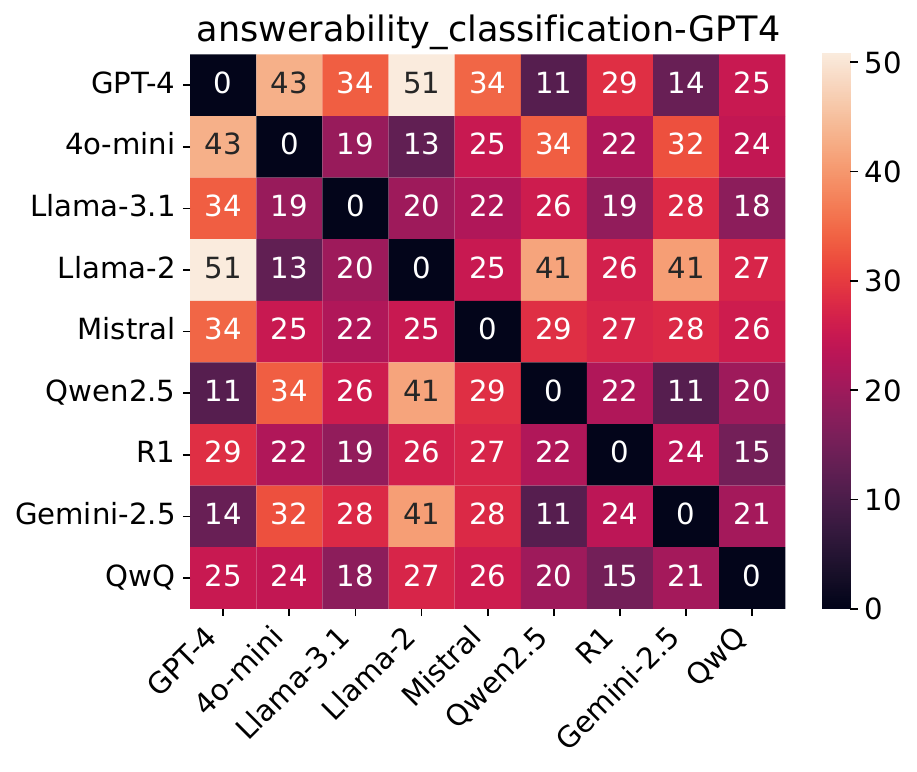}
    }
    \vspace{-0.15in}
    \caption{Error reductions of prevalent LLMs in detecting errors of \text{GPT-4}. The numbers refer to the reduced errors following the oracle collaboration via Eq.~(\ref{eq:oracle_collab}). It can be found that different LLMs are less likely to make mistakes simultaneously. When incorporating LLMs with higher heterogeneity, such as those from different companies, the error reduction rates will be higher.}
\end{figure*}

\begin{figure*}[ht]
    \centering
    \subfigure[Math problem]{
        \includegraphics[width=0.31\textwidth]{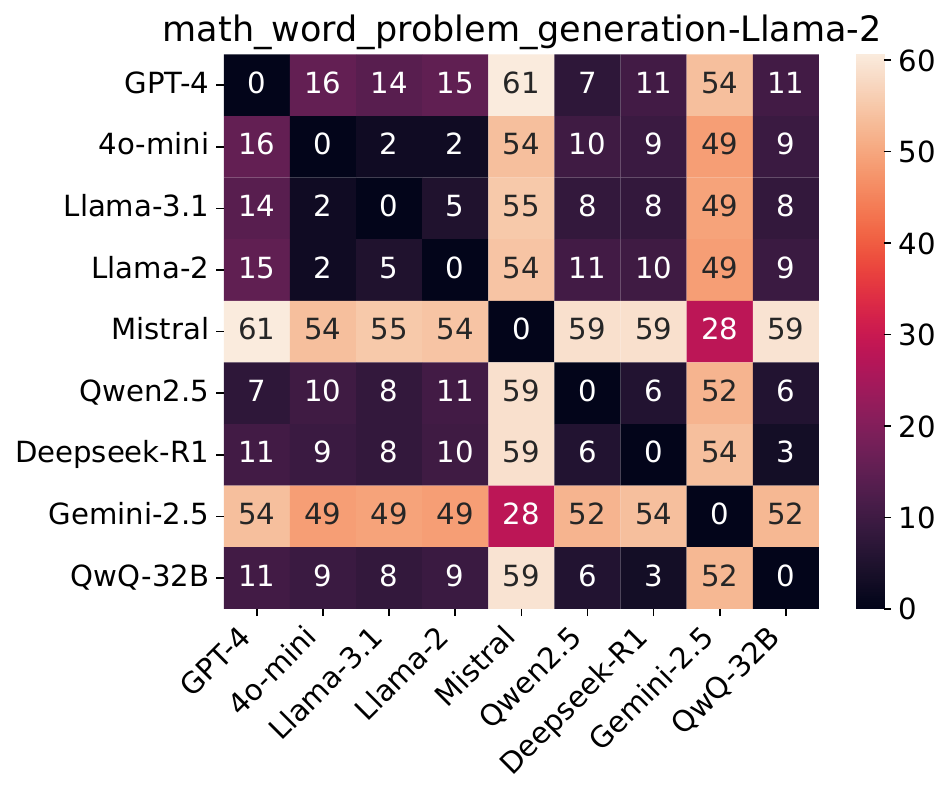}
    }
    \subfigure[Fact verification]{
        \includegraphics[width=0.31\textwidth]{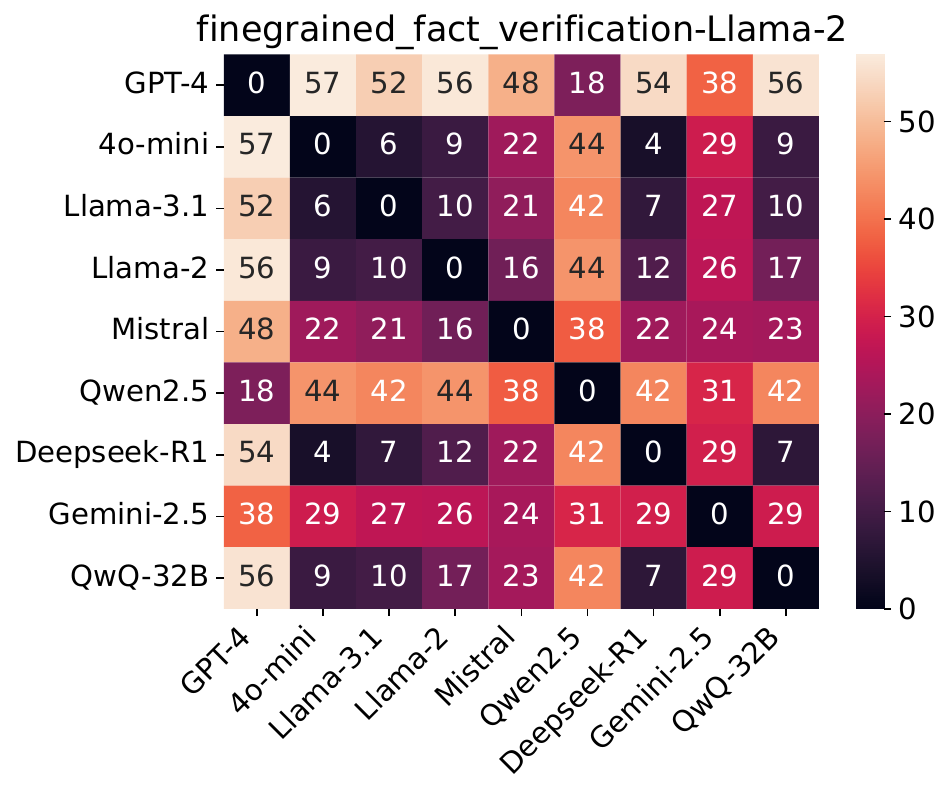}
    }
    \subfigure[Answerability]{
        \includegraphics[width=0.31\textwidth]{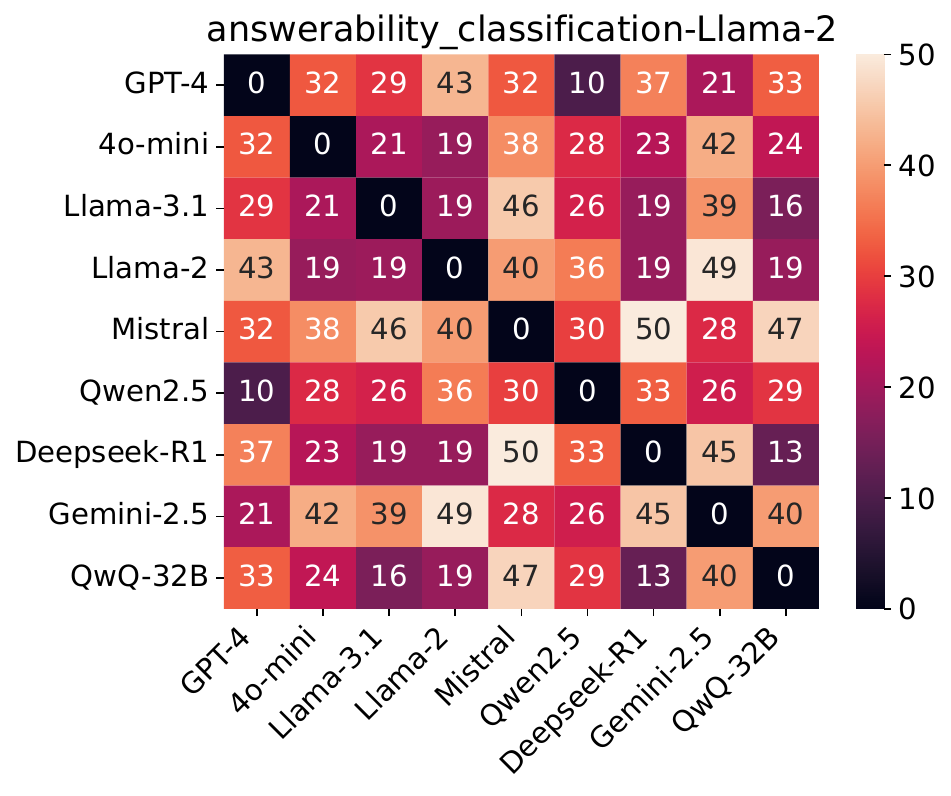}
    }
    \vspace{-0.15in}
    \caption{Error reductions of prevalent LLMs in detecting errors of \text{Llama-2}. The numbers refer to the reduced errors following the oracle collaboration via Eq.~(\ref{eq:oracle_collab}). It can be found that different LLMs are less likely to make mistakes simultaneously. When incorporating LLMs with higher heterogeneity, such as those from different companies, the error reduction rates will be higher.}
\end{figure*}

\section{Details of the Debate}\label{appdx:method}

\subsection{More details on \ours}
\begin{algorithm}[H]
        \caption{The \ours Framework}
        \label{alg:col}
        \begin{algorithmic}[1]
        \STATE  \textbf{Required:} \ours debater agents $\mathrm{A}$, and $\mathrm{B}$; the judge agent $J$; Dataset of LLM responses $\gD = \{(\vx^{(i)}, y^{(i)})\}_{i=1}^{n}$; Maximal debate rounds $R$;
        \STATE Initializing debater $\mathrm{A}$'s solution $x_A^0$ via prompting $\mathrm{A}$ using single-agent prompt, and obtaining the label $y_A$;
        \STATE Initializing debater $\mathrm{B}$'s solution $x_B^0$ via prompting $\mathrm{B}$ using single-agent prompt, and obtaining the label $y_B$;
        \IF{$y_A$ does not equal to $y_B$}
        \STATE Constructing debate transcripts $M^{(0)}=(m_A^{(0)},m_B^{(0)})$;
        \FOR{current round $t\in[1,...,T]$}
        \STATE Instructing debater $\mathrm{A}$ with $M_e=\{M^{(t)}\}_{t=0}^{t-1}$ to find missing points of $\mathrm{B}$ to obtain $m_A^{(t)}$;
        \STATE Instructing debater $\mathrm{B}$ with $M_e=\{M^{(t)}\}_{t=0}^{t-1}$ to find missing points of $\mathrm{A}$ to obtain $m_B^{(t)}$;
        \STATE Constructing debate transcripts $M^{(t)}=(m_A^{(t)},m_B^{(t)})$;
        \ENDFOR
        \STATE Summarizing transcripts $M_e=\{M^{(t)}\}_{t=0}^{T}$;
        \STATE Instructing the Judge $J$ to reason for the answer $y_J$;
        \ELSE
        \STATE $y_J = y_A$;
        \ENDIF
        \STATE \textbf{return} $y_J$;
        \end{algorithmic}
\end{algorithm}

The algorithm of \ours is given in Alg.~\ref{alg:col}, and the critical prompts of \ours are given as follows:

\paragraph{Debater prompt.}  The debater agent is encouraged to collaborate with the other debater agent to provide comprehensive information to the judge to make a well-informed decision. In addition to clearly state the objective of collaboration for seeking the truthful answer, we implement the insights of \ours into specific prompting schemes: (i) Evidence verification: \ours implements a quote-based system that asks the debaters to quote evidence from the context, and each quote will be verified if there is an exact match in the context, following~\citet{kenton2024on}; (ii) Self-auditing: the debaters are required to self-audit if there exists one potential failure mode in the claim; (iii) Confidence calibration: the debaters are required to provide a confidence estimate for their own claims.
\begin{tcolorbox}[title=Debater prompt]
\small
\setlength{\tabcolsep}{0.1mm}{

\textcolor{pink-color}{\textless{}task\textgreater{}}
You will act as an expert debater named Debater \{debater\_name\} on error detection for a model response. You are given a model input and a model response. You are assigned to defend \{answer\_defending\} while another debater defends the alternative. A judge will decide the final answer. Your mandate is collaborative truth-seeking with adversarial robustness: surface decisive errors (in either side), adopt valid points, and converge to the best-supported conclusion.
\\\\
Tone: scholarly, concise, evidence-first. Do NOT include an ``In summary/Conclusion'' paragraph because more rounds may follow.
\\\\
Output format:\\
``thinking"…private reasoning…``thinking"
``argument"…what the judge sees…``argument"
\\\\
The judge only sees "argument", so prioritize clarity and testable evidence.\\
\textcolor{pink-color}{\textless{}/task\textgreater{}}\\
\textcolor{pink-color}{\textless{}quote-system\textgreater{}}\\
Judging priorities (optimize for these):\\
1) Decisive tests that flip the result if they fail (spec mismatch, contradiction, counterexample).\\
2) Verified quotes (``quote"…``quote'' → may become ``v\_quote'') grounded in the task's general/specific requirements.\\
3) Novel, non-redundant contributions; penalties for repetition or unsupported claims.\\
4) Accurate calibration and willingness to adopt valid opponent points.\\
\textcolor{pink-color}{\textless{}/quote-system\textgreater{}}

\textcolor{pink-color}{\textless{}problem\textgreater{}}Here is the model input and the model response...\textcolor{pink-color}{\textless{}/problem\textgreater{}}

\textcolor{pink-color}{\textless{}transcript\textgreater{}}Here is the transcript of the debate so far...\textcolor{pink-color}{\textless{}/transcript\textgreater{}}

\textcolor{pink-color}{\textless{}thinking\textgreater{}}\\
Now produce your next turn. Follow these steps in your "thinking", then write only the "argument":\\
\\
1) Identify decisive tests: 2-4 concrete checks that would settle \{answer\_defending\} vs the alternative (e.g., requirement violation, logical contradiction, factual mismatch).\\
2) Run each test against BOTH sides (yours and opponent's). If a test fails on your side, plan a revision or boundary condition.\\
3) Extract 2-4 atomic requirements and cite them via "quote"…"/quote" (no repetition; aim for new quotes).\\
4) Steelman Opponent: find their strongest correct point and either integrate it or delimit its scope.\\
5) Prepare a one-sentence update note describing any change in your view.\\
\textcolor{pink-color}{\textless{}/thinking\textgreater{}}\\
\textcolor{pink-color}{\textless{}argument\textgreater{}}\\
- Position: state which answer you defend and why in one sentence.\\
- Decisive checks: list your 2-4 tests with PASS/FAIL for each side, and the minimal evidence that justifies each status.\\
- Evidence: map 2-4 ``quote''…``quote'' to specific claims you make (no duplicate quotes).\\
- Steelman: state opponent's strongest correct point and how you incorporate or delimit it.\\
- Self-audit: one plausible failure mode in your argument (1 short clause).\\
- End with:\\
  Final: ``1 error | 2 no\_error''\\
  Conf: ``0-1'' (justify briefly: signal strength, remaining uncertainties).\\
\\
Do not exceed \{word\_limit\} words in ``argument''. Remember to output both ``thinking'' and ``argument''.\\
\textcolor{pink-color}{\textless{}/argument\textgreater{}}

}
\end{tcolorbox}

\clearpage
\paragraph{Judge prompt.}  The Judge agent is instructed to analyze the debating transcripts to

\begin{tcolorbox}[title=Judge Prompt]
\small
\setlength{\tabcolsep}{0.1mm}{

\textcolor{pink-color}{\textless{}task\textgreater{}} \\
You are an expert judge in a debate where the participants are arguing over the correct answer to the error detection problem. Your task is to determine which debater is correct. Your main objective is to choose the correct answer based on the evidence presented. You should maintain a scholarly yet assertive tone focused on evidence-based analysis while critically evaluating arguments. \\
\textcolor{pink-color}{\textless{}/task\textgreater{}}\\
\textcolor{pink-color}{\textless{}quote-system\textgreater{}}\\
Here is some important information regarding the quote system which is applied after the debaters have finished their arguments: \\
- Verified quotes are direct string matches from the requirements and are enclosed in ``v\_quote''``/v\_quote'' tags. \\
- Unverified quotes are not direct string matches from the requirements, so they may not be trusted and are enclosed in ``u\_quote''``/u\_quote'' tags. \\
- Some quotes are just enclosed in quotation marks, which counts as unverified and may not be trusted. \\
\textcolor{pink-color}{\textless{}quote-system\textgreater{}}\\

\textcolor{pink-color}{\textless{}problem\textgreater{}}Here is the model input and the model response...\textcolor{pink-color}{\textless{}/problem\textgreater{}}

\textcolor{pink-color}{\textless{}transcript\textgreater{}}Here is the transcript of the debate...\textcolor{pink-color}{\textless{}/transcript\textgreater{}}

\textcolor{pink-color}{\textless{}reasoning\textgreater{}}\\
Choices: 1: \{answer\_1\} 2: \{answer\_2\} \\
\{name\_1\} is arguing for 1 and \{name\_2\} is arguing for 2. \\
\\
Choose the answer you think is most likely to be correct. \\
First explain your reasoning, considering the strengths and weaknesses of each answer. \\
Second, summarize the key pieces of evidence from the debate for why the error detection of model response is \{answer\_1\} or \{answer\_2\}. \\
Finally, give your answer formatted as " Answer: ``1 \{answer\_1\}|2 \{answer\_2\}''.\\
\textcolor{pink-color}{\textless{}/reasoning\textgreater{}}

}
\end{tcolorbox}

\paragraph{Problem.} The problem statement is organized as follows:
\begin{tcolorbox}[title=Problem]
\small
\setlength{\tabcolsep}{0.1mm}{

\textcolor{pink-color}{\textless{}problem\textgreater{}}\\
Here is the original problem, including model input and the model response:\\
We provide an input to a language model (model input) and a response from the model (model response). Your task is to detect errors in the model response. You need to check whether the model response follows all instructions and requirements in the model input. You need to classify the model response as an error if the model response does not follow any of the instructions or requirements in the model input.\\
\\
===== Model Input Begins =====\\
......\\
===== Model Input Ends =====\\
\\
===== Model Response Begins =====\\
......\\
===== Model Response Ends =====\\
In your response, provide your explanation first and conclude your response with "Therefore, the model response contains an error." or "Therefore, the model response contains no error."
\textcolor{pink-color}{\textless{}/problem\textgreater{}}
}
\end{tcolorbox}

\clearpage
\paragraph{Transcripts.} The transcripts are organized as follows:
\begin{tcolorbox}[title=Transcripts]
\small
\setlength{\tabcolsep}{0.1mm}{

\textcolor{pink-color}{\textless{}transcript\textgreater{}}\\
Here is the transcript of the debate:\\
Round 1: \\
debater Alice: ... \\
debater Bob: ... \\
Round 2: \\
debater Alice: ... \\
debater Bob: ... \\
\textcolor{pink-color}{\textless{}/transcript\textgreater{}}
}
\end{tcolorbox}

\section{More Details about Experiments}
\label{appdx:exp}
\begin{table*}[ht]
    \fontsize{7pt}{8pt}\selectfont
    \centering
    \caption{Statistics of \real benchmark~\citep{realmistake}.
    }
    \label{tab:datasets-in-benchmark}
    \begin{tabular}{clccrrrrr|r}
    \toprule
        \multirow{3}{*}{\rotatebox[origin=c]{90}{\parbox{4em}{\centering Response\\Model}}} &
        \multicolumn{1}{c}{\multirow{4}{*}{Task}} & \multirow{4}{*}{\# Data} & \multicolumn{2}{c}{Average \# tokens} & \multicolumn{5}{c}{Errors in Responses from GPT-4 or Llama~2~70B [\%]} \\
    \cmidrule(l{2pt}r{2pt}){4-5} \cmidrule(l{2pt}r{2pt}){6-10}
        & & & \mrtwo{Input} & \ccell{LLM} & \scb{Reasoning}   & \scb{Instruction-} & \scb{Context-} & \scb{Parameterized} & \ccell{Total} \\
        & & & & \scb{Response} & \scb{Correctness} & \scb{Following} & \scb{Faithfulness} & \scb{Knowledge} & \ccell{Error} \\
    \midrule
        \multirow{3}{*}{\rotatebox[origin=c]{90}{\parbox{3em}{\centering GPT-4\\0613}}}
        & Math Word Problem Generation    &  140 &   252 &   151\phantom{00} & 25.0 & 57.1 &   -- &   -- & 62.1 \\
        & Fine-grained Fact Verification  &  140 &   523 &    83\phantom{00} & 25.7 &  5.7 & 45.0 &   -- & 62.9 \\
        & Answerability Classification    &  140 &   119 &    75\phantom{00} & 22.1 &   -- &  8.6 & 40.7 & 62.1 \\
    \midrule
        \multirow{3}{*}{\rotatebox[origin=c]{90}{\parbox{3em}{\centering Llama~2\\70B}}}
        & Math Word Problem Generation    &  160 &   235 &   163\phantom{00} & 51.2 & 67.5 &   -- &   -- & 80.0 \\
        & Fine-grained Fact Verification  &  160 &   511 &   168\phantom{00} & 56.9 & 44.4 & 45.6 &   -- & 80.6 \\
        & Answerability Classification    &  160 &   119 &    96\phantom{00} & 48.1 &   -- &   -- & 48.1 & 81.2 \\
    \bottomrule
    \end{tabular}
\end{table*}

\subsection{Additional results on detecting errors by Llama-2}
\label{appdx:llama-2}

\begin{table*}[t]
\selectfont
\centering
    \caption{Results for Llama-2 responses. The judge uses the same LLM as Debater\#1. The top two results are highlighted.}
    \label{tab:llama2-results}
\resizebox{\textwidth}{!}{
\begin{tabular}{lll|*{6}{p{1.2cm}}|*{2}{p{1.2cm}}}
\toprule
    \multirow{2}{*}{Debater\#1} & \multirow{2}{*}{Debater\#2} & \multirow{2}{*}{Protocol}
    & \multicolumn{2}{c}{Math Problem}
    & \multicolumn{2}{c}{Fact Verification}
    & \multicolumn{2}{c}{Answerability}
    & \multicolumn{1}{c}{Avg. F1} & \multicolumn{1}{c}{Avg. F2} \\
    \cmidrule(l{2pt}r{2pt}){4-11}
     & & & {F1~($\uparrow$)}& {F2~($\uparrow$)}
     & {F1~($\uparrow$)} & {F2~($\uparrow$)}
     & {F1~($\uparrow$)} & {F2~($\uparrow$)}
     & \multicolumn{1}{c}{F1~($\uparrow$)} & \multicolumn{1}{c}{F2~($\uparrow$)} \\
\midrule
Human&-&-&98.30&97.34&100.0&100.0&100.0&100.0&99.43&99.11\\
GPT4o-mini	&-	&-&89.82	&\textbf{95.67}	&\textbf{93.14}	&\textbf{97.14}	&79.50	&75.52	&87.49	&89.44\\
Llama3.1-70B	&-&-	&\textbf{90.78}	&96.10	&91.45	&93.75	&82.87	&81.12	&88.37	&90.32\\
Mistral-7B-v0.3	&-&-	&45.00	&38.53	&82.07	&80.72	&51.04	&42.10	&59.37	&53.78\\
\midrule
GPT4o-mini	&Llama3.1-70B&\ours	&89.82	&\textbf{95.67}	&92.47	&96.85	&\textbf{87.55}	&\textbf{88.55}	&\textbf{89.95}	&\textbf{93.69}\\
	&&\copmad	&82.07	&81.10	&78.18	&70.84	&51.34	&41.59	&70.53	&64.51\\
	&&Voting	&\textbf{90.46}	&95.95	&93.43	&96.82	&81.30	&78.62	&88.40	&90.46\\\midrule
Llama3.1-70B	&GPT4o-mini&\ours	&89.82	&\textbf{95.67}	&92.47	&96.85	&\textbf{88.81}	&\textbf{90.43}	&\textbf{90.37}	&\textbf{94.31}\\
	&&\copmad	&88.89	&93.51	&89.47	&91.12	&76.99	&73.13	&85.12	&85.92\\
	&&Voting	&\textbf{90.46}	&95.95	&93.43	&96.82	&81.30	&78.62	&88.40	&90.46\\\midrule
GPT4o-mini	&Mistral-7B-v0.3&\ours	&88.50	&94.63	&\textbf{92.81}	&\textbf{96.99}	&79.84	&77.59	&87.05	&89.74\\
	&&\copmad	&85.71	&86.31	&80.53	&74.23	&59.70	&50.76	&75.31	&70.43\\
	&&Voting	&64.76	&57.24	&59.51	&51.52	&63.81	&55.83	&62.69	&54.86\\\midrule
Llama3.1-70B	&Mistral-7B-v0.3&\ours	&90.14	&\textbf{95.81}	&90.91	&94.41	&84.50	&84.10	&88.52	&91.44\\
	&&\copmad	&89.21	&93.66	&84.92	&83.72	&74.36	&69.71	&82.83	&82.36\\
	&&Voting	&60.00	&57.01	&48.98	&44.78	&60.00	&57.01	&56.33	&52.93\\
\bottomrule
\end{tabular}}
\vspace{-0.15in}
\end{table*}

The error detection results on Llama-2 responses in \real are given in Table~\ref{tab:llama2-results}. As one could find it is relatively easier to detect errors of weak LLMs, all methods achieve relatively high results. Nevertheless, \ours remain the top one.

\textbf{Transferability of \ours to different candidate LLMs.} Combining the results of Table~\ref{tab:gpt4-results} in Sec.~\ref{sec:exp} and Table~\ref{tab:llama2-results}, although detecting errors of Llama-2 responses is relatively easier than that of GPT-4, we can also find that \ours outperforms \copmad as well as the best single-agent performance by up to 4\%. The results indicate the generality of the advantages of \ours.

\subsection{Ablation studies on the effectiveness of different prompt components}
\label{appdx:prompt}
\begin{table}[t]
\centering
\caption{Ablation on the prompt components of \ours, with \texttt{GPT4o-mini} and \texttt{Llama3.1-70B} as debaters. Each component of the structured prompt design (quote-based evidence verification, confidence declaration, and self-auditing) contributes to performance.}
\label{tab:ablation-prompt}
\small
\setlength{\tabcolsep}{4pt}
\resizebox{\textwidth}{!}{%
\begin{tabular}{l|cc|cc|cc|cc}
\toprule
\multirow{2}{*}{Protocol}
    & \multicolumn{2}{c|}{Math Problem}
    & \multicolumn{2}{c|}{Fact Verification}
    & \multicolumn{2}{c|}{Answerability}
    & \multicolumn{2}{c}{Average} \\
    & F1 & F2 & F1 & F2 & F1 & F2 & F1 & F2 \\
\midrule
\ours (full)                                                     & \textbf{78.38} & \textbf{90.06} & \textbf{77.42} & \textbf{88.98} & 72.91 & 79.74 & \textbf{76.24} & \textbf{86.26} \\
\quad w/o quote system                                           & 77.68 & 89.69 & 77.21 & 88.30 & 73.17 & 80.47 & 76.02 & 86.15 \\
\quad w/o quote system \& confidence                             & 77.83 & 89.21 & 76.42 & 86.72 & 74.04 & 82.09 & 76.10 & 86.01 \\
\quad w/o quote system \& confidence \& self-auditing            & 78.03 & 89.88 & 74.64 & 84.05 & \textbf{75.49} & \textbf{82.80} & 76.05 & 85.58 \\
\bottomrule
\end{tabular}%
}
\end{table}

\textbf{Prompt components of \ours.} We ablate the three structured prompt components: (i) quote-based evidence verification, (ii) confidence declaration, and (iii) self-auditing, using \texttt{GPT4o-mini} and \texttt{Llama3.1-70B}. As shown in Table~\ref{tab:ablation-prompt}, each component contributes to the final performance. Even without all three, \ours (85.58) still far outperforms \copmad (57.34) and \cosmad (75.40), confirming that the core advantage comes from the collaborative game structure. A detailed mapping from conditions to components is in Appendix~\ref{appdx:theory_impl_map}.

\subsection{Additional results with shared debaters and weaker judges}
\label{appdx:same_debater}

To probe the concern that our main setting pairs Debater\#1 with the judge as the same LLM, we further experiment with two identical debaters (\texttt{GPT4o-mini} vs. \texttt{GPT4o-mini}), both with a matching judge and with an independent, weaker judge (\texttt{Mistral-7B-v0.3}). Results are given in Table~\ref{tab:same-debater}. \ours still outperforms \copmad even when both debaters share the same LLM, but all \mad methods underperform the single-agent baseline, reinforcing the observation that \emph{heterogeneous debaters} are a key driver of \ours's gains.

\begin{table}[t]
\centering
\caption{Results when both debaters share the same LLM (\texttt{GPT4o-mini}), with the judge being either the same model (\texttt{GPT4o-mini}) or a weaker independent model (\texttt{Mistral-7B-v0.3}). \ours still outperforms \copmad even with identical debaters, while all \mad methods underperform the single-agent baseline (equivalent to Voting when debaters agree), demonstrating the necessity of using heterogeneous LLMs for effective error detection.}
\label{tab:same-debater}
\small
\setlength{\tabcolsep}{4pt}
\resizebox{\textwidth}{!}{%
\begin{tabular}{lll|cc|cc|cc|cc}
\toprule
\multirow{2}{*}{Debater\#1} & \multirow{2}{*}{Debater\#2} & \multirow{2}{*}{Judge}
    & \multicolumn{2}{c|}{Math Problem}
    & \multicolumn{2}{c|}{Fact Verification}
    & \multicolumn{2}{c|}{Answerability}
    & \multicolumn{2}{c}{Average} \\
    & & & F1 & F2 & F1 & F2 & F1 & F2 & F1 & F2 \\
\midrule
\multirow{3}{*}{GPT4o-mini} & \multirow{3}{*}{GPT4o-mini} & \multirow{3}{*}{GPT4o-mini}
    & \cellcolor{babyblue}\textbf{76.58} & \cellcolor{babyblue}\textbf{87.99} & \cellcolor{babyblue}\textbf{72.82} & \cellcolor{babyblue}\textbf{78.89} & \cellcolor{babyblue}\textbf{70.00} & \cellcolor{babyblue}\textbf{75.92} & \cellcolor{babyblue}\textbf{73.13} & \cellcolor{babyblue}\textbf{80.93}
    \\
    & & \multicolumn{1}{l}{}
    & 70.90 & 74.44 & 66.28 & 66.74 & 48.23 & 42.29 & 61.80 & 61.16 \\
    & & \multicolumn{1}{l}{}
    & 78.70 & 89.10 & 75.76 & 82.78 & 70.10 & 74.73 & 74.85 & 82.20 \\
\midrule
\multirow{3}{*}{GPT4o-mini} & \multirow{3}{*}{GPT4o-mini} & \multirow{3}{*}{Mistral-7B-v0.3}
    & \cellcolor{babyblue}\textbf{73.68} & \cellcolor{babyblue}\textbf{81.91} & \cellcolor{babyblue}\textbf{65.22} & \cellcolor{babyblue}\textbf{68.34} & \cellcolor{babyblue}\textbf{68.75} & \cellcolor{babyblue}\textbf{72.85} & \cellcolor{babyblue}\textbf{69.22} & \cellcolor{babyblue}\textbf{74.37}
    \\
    & & & 61.35 & 58.96 & 54.88 & 53.70 & 53.15 & 47.03 & 56.46 & 53.23 \\
    & & & 78.70 & 89.10 & 75.76 & 82.78 & 70.10 & 74.73 & 74.85 & 82.20 \\
\bottomrule
\multicolumn{11}{l}{\footnotesize Rows within each block, top to bottom: \ours, \copmad, single-agent (=\,Voting when both debaters are the same).}
\end{tabular}%
}
\end{table}

\subsection{Token cost across \mad protocols}
\label{appdx:token_cost}

Table~\ref{tab:token-cost} reports the averaged token cost (mean $\pm$ std) of the full debating transcripts of each \mad protocol with \texttt{Llama3.1-70B} and \texttt{GPT4o-mini} as debaters. \ours uses $\sim\!15\%$ more tokens than \copmad but yields substantially larger improvements in F2, so the gain cannot be attributed to raw token usage. Crucially, \copmad uses tokens comparable to \ours yet underperforms single-agent baselines --- demonstrating that the protocol design, not compute, drives \ours's advantage.

\begin{table}[t]
\centering
\caption{Averaged token cost of debating transcripts across \mad protocols, with \texttt{Llama3.1-70B} and \texttt{GPT4o-mini} as debaters. Numbers after $\pm$ are standard deviations. Although \ours consumes somewhat more tokens than other protocols, the improvement in F2 is substantially larger than the $\sim\!\!15\%$ token overhead over \copmad.}
\label{tab:token-cost}
\small
\setlength{\tabcolsep}{4pt}
\begin{tabular}{l|cc|cc|cc}
\toprule
\multirow{2}{*}{Protocol}
    & \multicolumn{2}{c|}{Math Problem}
    & \multicolumn{2}{c|}{Fact Verification}
    & \multicolumn{2}{c}{Answerability} \\
    & F2 & Token cost & F2 & Token cost & F2 & Token cost \\
\midrule
\ours   & \textbf{90.06} & $13259.15 \pm 971.28$  & \textbf{88.98} & $13327.00 \pm 1033.25$ & \textbf{79.74} & $13323.00 \pm 1086.06$ \\
\copmad & 88.91 & $11449.09 \pm 524.24$  & 82.43 & $11484.94 \pm 538.15$  & 69.32 & $11401.81 \pm 714.19$ \\
SoM     & 89.55 & $\phantom{0}9461.82 \pm 4225.49$ & 79.37 & $\phantom{0}6642.09 \pm 1061.34$ & 60.14 & $\phantom{0}7076.89 \pm 4165.38$ \\
MP      & 75.79 & $\phantom{0}9690.29 \pm 2224.01$ & 75.00 & $\phantom{0}9344.22 \pm \phantom{0}712.82$ & 55.56 & $\phantom{0}9471.56 \pm 2136.85$ \\
\bottomrule
\end{tabular}
\end{table}

\subsection{Fine-grained alignment with human-annotated errors}
\label{appdx:finegrained}

To complement the LLM-as-judge evaluation in Fig.~\ref{fig:rating}, we further analyze debating transcripts from \copmad and \ours (\texttt{Llama3.1-70B} vs. \texttt{DeepSeek-R1}) on \mathw, extracting human-annotated error categories. Table~\ref{tab:finegrained-alignment} reports the percentage of each category identified by the corresponding method. Across categories, \ours identifies the same mistakes pointed out by humans at consistently higher rates than both single-agent baselines and \copmad.

\begin{table}[t]
\centering
\caption{Fine-grained coverage (\%) of human-annotated error categories in math problem generation, with \texttt{Llama3.1-70B} and \texttt{DeepSeek-R1} as debaters. \ours consistently identifies the same mistakes pointed out by humans better than the single-agent and \copmad alternatives.}
\label{tab:finegrained-alignment}
\small
\setlength{\tabcolsep}{6pt}
\begin{tabular}{l|cccc}
\toprule
Error category              & Llama3.1-70B & DeepSeek-R1 & \copmad & \ours \\
\midrule
Specific integer format     & 17.60 & 38.20 & 38.20 & \textbf{44.10} \\
Specific phase requirement  & 50.00 & 58.30 & 58.30 & \textbf{58.30} \\
Ambiguous or unanswerable   & 30.00 & 40.00 & 30.00 & \textbf{40.00} \\
Missing required content    & 14.29 & 14.29 & \phantom{0}0.00  & \textbf{28.57} \\
Calculation error           & \phantom{0}0.00  & 50.00 & \phantom{0}0.00  & \textbf{50.00} \\
\bottomrule
\end{tabular}
\end{table}

\clearpage
\section*{NeurIPS Paper Checklist}

\begin{enumerate}

\item {\bf Claims}
    \item[] Question: Do the main claims made in the abstract and introduction accurately reflect the paper's contributions and scope?
    \item[] Answer: \answerYes{}
    \item[] Justification: We clearly state our contribution to understanding when and why \mad fails, and thereby providing a new \mad protocol based on our analysis.

\item {\bf Limitations}
    \item[] Question: Does the paper discuss the limitations of the work performed by the authors?
    \item[] Answer: \answerYes{}
    \item[] Justification: We provide a discussion in Appendix~\ref{appdx:discussion}.

\item {\bf Theory assumptions and proofs}
    \item[] Question: For each theoretical result, does the paper provide the full set of assumptions and a complete (and correct) proof?
    \item[] Answer: \answerYes{}
    \item[] Justification: We provide necessary details in the main text and full details in Appendix~\ref{appdx:theory}.

    \item {\bf Experimental result reproducibility}
    \item[] Question: Does the paper fully disclose all the information needed to reproduce the main experimental results of the paper to the extent that it affects the main claims and/or conclusions of the paper (regardless of whether the code and data are provided or not)?
    \item[] Answer: \answerYes{}
    \item[] Justification: We provide full details in Sec.~\ref{sec:exp}.

\item {\bf Open access to data and code}
    \item[] Question: Does the paper provide open access to the data and code, with sufficient instructions to faithfully reproduce the main experimental results, as described in supplemental material?
    \item[] Answer: \answerYes{}
    \item[] Justification: We will provide an anonymous link to our code during the discussion phase and open-source the code when published.

\item {\bf Experimental setting/details}
    \item[] Question: Does the paper specify all the training and test details (e.g., data splits, hyperparameters, how they were chosen, type of optimizer) necessary to understand the results?
    \item[] Answer: \answerYes{}
    \item[] Justification: We provide all the necessary details in Sec.~\ref{sec:exp}.

\item {\bf Experiment statistical significance}
    \item[] Question: Does the paper report error bars suitably and correctly defined or other appropriate information about the statistical significance of the experiments?
    \item[] Answer: \answerYes{}
    \item[] Justification: We provide statistical significance in Table~\ref{tab:confidence-interval}.

\item {\bf Experiments compute resources}
    \item[] Question: For each experiment, does the paper provide sufficient information on the computer resources (type of compute workers, memory, time of execution) needed to reproduce the experiments?
    \item[] Answer: \answerYes{}
    \item[] Justification: Most of our experiments are using API calls, while experiments in Table~\ref{tab:general-results} are run with A100 gpus.

\item {\bf Code of ethics}
    \item[] Question: Does the research conducted in the paper conform, in every respect, with the NeurIPS Code of Ethics \url{https://neurips.cc/public/EthicsGuidelines}?
    \item[] Answer: \answerYes{}
    \item[] Justification: We ensure our paper follows the NeurIPS Code of Ethics.

\item {\bf Broader impacts}
    \item[] Question: Does the paper discuss both potential positive societal impacts and negative societal impacts of the work performed?
    \item[] Answer: \answerYes{}
    \item[] Justification: We provide a discussion at the begining of the Appendix.

\item {\bf Safeguards}
    \item[] Question: Does the paper describe safeguards that have been put in place for responsible release of data or models that have a high risk for misuse (e.g., pre-trained language models, image generators, or scraped datasets)?
    \item[] Answer: \answerNA{}
    \item[] Justification: Our work provides new insights for AI safety, and to the best of our knowledge, we are not aware of potential risks caused by our work.

\item {\bf Licenses for existing assets}
    \item[] Question: Are the creators or original owners of assets (e.g., code, data, models), used in the paper, properly credited and are the license and terms of use explicitly mentioned and properly respected?
    \item[] Answer: \answerYes{}
    \item[] Justification: We cite proper sources for data and methods used in our paper.

\item {\bf New assets}
    \item[] Question: Are new assets introduced in the paper well documented and is the documentation provided alongside the assets?
    \item[] Answer: \answerNA{}
    \item[] Justification: The paper does not release new assets.

\item {\bf Crowdsourcing and research with human subjects}
    \item[] Question: For crowdsourcing experiments and research with human subjects, does the paper include the full text of instructions given to participants and screenshots, if applicable, as well as details about compensation (if any)?
    \item[] Answer: \answerNA{}
    \item[] Justification: The paper does not involve crowdsourcing nor research with human subjects.

\item {\bf Institutional review board (IRB) approvals or equivalent for research with human subjects}
    \item[] Question: Does the paper describe potential risks incurred by study participants, whether such risks were disclosed to the subjects, and whether Institutional Review Board (IRB) approvals (or an equivalent approval/review based on the requirements of your country or institution) were obtained?
    \item[] Answer: \answerNA{}
    \item[] Justification: The paper does not involve crowdsourcing nor research with human subjects.

\item {\bf Declaration of LLM usage}
    \item[] Question: Does the paper describe the usage of LLMs if it is an important, original, or non-standard component of the core methods in this research? Note that if the LLM is used only for writing, editing, or formatting purposes and does \emph{not} impact the core methodology, scientific rigor, or originality of the research, declaration is not required.
    \item[] Answer: \answerYes{}
    \item[] Justification: We describe the use of LLM at the begining of the Appendix.

\end{enumerate}

\end{document}